\ifdefined\pdfminorversion
\fi
\PassOptionsToPackage{table}{xcolor}
\documentclass[11pt, a4paper, logo, copyright, nonumbering]{map}
\usepackage{CJKutf8}
\usepackage{xargs}  
\usepackage{todonotes}
\usepackage{amsmath}
\usepackage{dsfont}

\usepackage{longtable}

\usepackage{svg}
\usepackage{fontawesome}
\usepackage{array}
\usepackage{tabularx}
\usepackage{latexsym}
\usepackage{graphicx}
\usepackage[utf8]{inputenc} % allow utf-8 input
\usepackage[utf8]{inputenc} % allow utf-8 input
\usepackage{amssymb} % for math symbols like \geq
\usepackage{url}            % simple URL typesetting
\usepackage{amsfonts}       % blackboard math symbols
\usepackage{nicefrac}       % compact symbols for 1/2, etc.
\usepackage{microtype}      % microtypography
\usepackage{xspace}

\usepackage{listings}
\usepackage{makecell}
\usepackage{inconsolata}
\usepackage{multicol}
\usepackage{amstext}

\definecolor{darkblue}{RGB}{84, 112, 198}

\definecolor{lightblue}{rgb}{0.85, 0.95, 1.0}    % 浅蓝
\definecolor{lightgreen}{rgb}{0.90, 1.0, 0.90}    % 浅绿
\definecolor{lightorange}{rgb}{1.0, 0.95, 0.85}   % 浅橙
\definecolor{lightpurple}{rgb}{0.95, 0.90, 1.0}   % 浅紫
\definecolor{lightgray}{rgb}{0.97, 0.97, 0.97}    % 浅灰

\usepackage{tikz}
\usetikzlibrary{calc}
\definecolor{battery-empty}{rgb}{0.9, 0.9, 0.9}
\newcommand{\difficultybar}[1]{%
  \begin{tikzpicture}[baseline, scale=0.5, every node/.style={scale=0.8}]
    \foreach \i in {1,2,3,4,5} {
      \ifnum\i>#1
        \draw[fill=battery-empty] (\i*0.5-0.5, 0) rectangle (\i*0.5, 0.25);
      \else
        \pgfmathsetmacro{\colorlevel}{80 - 12*(\i)} % i=1→54, i=2→48, i=3→42, i=4→36, i=5→30
        \edef\x{\noexpand\draw[fill=blue!\colorlevel!white, opacity=0.9] (\i*0.5-0.5, 0) rectangle (\i*0.5, 0.25);}
        \x
        \draw[blue!50!black] (\i*0.5-0.5, 0) rectangle (\i*0.5, 0.25);
      \fi
    }
    \fill[battery-empty!70] (2.5, 0.08) rectangle (2.6, 0.17);
    \draw[battery-empty!70!black] (2.5, 0.08) rectangle (2.6, 0.17);
  \end{tikzpicture}%
}
\renewcommand{\arraystretch}{0.96}

\definecolor{hidden-draw}{RGB}{20,68,106}
\definecolor{hidden-pink}{RGB}{255,245,247}
\usepackage[edges]{forest}

\usepackage{bm}
\usepackage{graphicx}
\usepackage{booktabs}
\usepackage{enumitem}
\usepackage{booktabs} 
\usepackage{array} 
\usepackage{xcolor}
\usepackage{soul}
\usepackage{makecell}
\usepackage{natbib}
\usepackage{CJKutf8}
\usepackage{xargs}
\usepackage{todonotes}
\usepackage{multirow}
\usepackage{cleveref}
\usepackage{subcaption}
\usepackage{url}
\usepackage{svg}
\usepackage{fontawesome}
\usepackage{xcolor}
\usepackage{float}
\usepackage{array}
\usepackage{ragged2e}
\usepackage[utf8]{inputenc}
\usepackage{booktabs}
\usepackage{array}
\usepackage{tabularx}
\usepackage{xcolor,colortbl}  % 引入xcolor包
\definecolor{boxcolor}{HTML}{d92523} % 框的颜色 #dea3a2
\definecolor{bulbcolor}{HTML}{e3b87f} % 灯泡的颜色 #e3b87f
\usepackage{url}
\usepackage{ragged2e}   % provides \RaggedRight for tabularx columns
\usepackage{makecell} 
\DeclareRobustCommand{\cmark}{\textcolor{green!60!black}{\ding{51}}} % ✓
\DeclareRobustCommand{\xmark}{\textcolor{red!70!black}{\ding{55}}}   % ✗

\usepackage{hyperref}       % hyperlinks
\usepackage{url}            % simple URL typesetting
\usepackage{booktabs}       % professional-quality tables
\usepackage{nicefrac}       % compact symbols for 1/2, etc.
\usepackage{microtype}      % microtypography
\usepackage{xcolor}         % colors
\usepackage{graphicx}
\usepackage{longtable}
\usepackage{array}
\usepackage{pbox}
\usepackage{tabularx}
\usepackage{multirow}
\usepackage{wrapfig}
\usepackage{pifont}
\usepackage{lipsum}
\usepackage{subcaption}
\usepackage{enumitem}
\usepackage{tikz}
\usepackage{xstring}

\usepackage{color-edits}

\usepackage{booktabs}
\usepackage{multirow}
\usepackage[table]{xcolor}
\usepackage{tabularx}

\usepackage{parskip} %去掉 paragraph indent

\usepackage{threeparttable} % Required for table notes
\usepackage{tabularx}
\usepackage{array}
\usepackage{ragged2e}
\usepackage{booktabs}

\newcolumntype{Y}{>{\RaggedRight\arraybackslash}X}
\usepackage{algorithm}
\usepackage{algpseudocode}

\usepackage{xspace}
\usepackage{times}
\usepackage{latexsym}
\usepackage{booktabs}
\usepackage{fvextra}
\usepackage{graphicx}
\usepackage{subcaption} % 需要加载 subcaption 宏包
\usepackage[T1]{fontenc}
\usepackage{xcolor}
\usepackage[utf8]{inputenc}

\usepackage{microtype}

\usepackage{inconsolata}
\usepackage{amsmath}
\usepackage{booktabs}
\usepackage{graphicx}
\usepackage{pifont}
\usepackage{multirow}          % For \multirow in tables
\usepackage{colortbl}          % Optional: improves table coloring support
\usepackage{algorithm}
\usepackage{algpseudocode}
\usepackage{xcolor}

\usepackage{amssymb}
\usepackage{amsthm}

 \usepackage{amsmath,amssymb}

\usepackage{alltt}
\usepackage[most,skins,theorems]{tcolorbox}
\tcbuselibrary{skins,breakable,fitting,hooks}
\usepackage{xcolor}

\hypersetup{
    colorlinks=true,            %链接颜色
    linkcolor=blue,             %内部链接
    filecolor=magenta,          %本地文档
    urlcolor=cyan,              %网址链接
    citecolor=purple,             % 引用颜色
    pdftitle={WorkFactory: Turning Professional Intents into Executable Work for Scalable Agent Training},
    pdfpagemode=FullScreen,
}

\usepackage{hyperref}
\usepackage{url}
\usepackage{booktabs}
\usepackage{array}
\usepackage{graphicx}
\usepackage[table]{xcolor}
\usepackage{multirow}

\usepackage{hyperref}
\usepackage{url}
\usepackage{enumitem}
\usepackage{booktabs}
\usepackage{array}
\usepackage{multirow}
\usepackage{graphicx}
\usepackage[table]{xcolor}
\usepackage{listings}
\usepackage{amsmath}
\usepackage{amsthm}
\usepackage{wrapfig}
\usepackage{needspace}
\usepackage{subcaption}

\theoremstyle{plain}

\definecolor{iquestblue}{HTML}{173C7F}
\definecolor{iquestazure}{HTML}{528FCC}

\newcommandx{\info}[2][1=]{\todo[linecolor=red,backgroundcolor=red!25,bordercolor=red,#1]{#2}}

\title{
\vspace{-0.2in}
\centering \fontsize{15pt}{16pt}\selectfont
WorkFactory: Turning Professional Intents into Executable Work \\for Scalable Agent Training

\vspace{-0.2in}
}
\author{
Ziyang Zeng\textsuperscript{1*},
Tingyang Chen\textsuperscript{2*},
Jia Deng\textsuperscript{3*},
Qiyao Wang\textsuperscript{4*},\\
\normalfont
Shuo Tang\textsuperscript{5*},
Yike Yang\textsuperscript{3,*},
Tao Ren\textsuperscript{6,*},
Wayne Xin Zhao\textsuperscript{$3\dagger$},\\
\normalfont
Feng Chang\textsuperscript{$1\dagger\ddagger$},
Jujie He\textsuperscript{1},
Chuan Hao\textsuperscript{1},
Ran Tao\textsuperscript{1},
Bryan Dai\textsuperscript{1}\\
\normalfont
\textsuperscript{1}IQuest Research\quad
\textsuperscript{2}Zhejiang University\\
\normalfont
\textsuperscript{3}Gaoling School of Artificial Intelligence, Renmin University of China\\
\normalfont
\textsuperscript{4}Shenzhen Institutes of Advanced Technology\\
\normalfont
\textsuperscript{5}Shanghai Jiao Tong University\quad
\textsuperscript{6}Peking University\\
\normalfont
\normalsize{
\textsuperscript{*}Equal contributors\quad
\textsuperscript{$\dagger$}Corresponding Authors\quad
\textsuperscript{$\ddagger$}Project Leader\\
\normalfont
Email: \texttt{yling@iquestlab.com},
\texttt{chenty@zju.edu.cn},\\
\normalfont
\texttt{batmanfly@gmail.com},
\texttt{fchang@iquestlab.com}
}
}

\newcommand{\ignore}[1]{}

\DefineVerbatimEnvironment{promptbox}{Verbatim}{%
  fontsize=\scriptsize,
  frame=single,
  framesep=3pt,
  breaklines=true,
  breakanywhere=true,
  breaksymbolleft={},
  breakindent=0pt
}
\usepackage[most]{tcolorbox}
\definecolor{darkorange}{RGB}{255, 140, 0}
\definecolor{lightgreen}{RGB}{145, 204, 117}
\definecolor{lightyellow}{RGB}{250, 200, 88}
\definecolor{lightred}{RGB}{238, 102, 102}
\definecolor{lightblue}{RGB}{115, 192, 222}

\definecolor{gray_1}{HTML}{B7B7B7}
\definecolor{gray_2}{HTML}{F0F0F0} 
\definecolor{frame_blue}{HTML}{A9D18E}

\newtcolorbox[auto counter, number within=section]{PromptBox}[2][]{
    enhanced,
    breakable,
    colback=gray_2, % 应用新的背景色
    colframe=gray_1,
    coltitle=white,
    fontupper=\small,
    fonttitle=\bfseries,
    title={#2}, 
    label={#1},
    arc=2pt,
    boxrule=1pt,
    left=2mm, right=2mm, top=2mm, bottom=2mm,
}

\usepackage{etoolbox}

\begin{abstract}
Scaling supervised fine-tuning for working agents requires not only broad coverage of professional tasks, but also expert demonstrations that faithfully carry out professional workflows to completion.
However, missing or inconsistent task information can disrupt expert demonstrations, weakening the resulting supervision.
We introduce WorkFactory, a scalable task-construction framework that turns professional intents into executable task environments through hybrid intent sourcing, semantic workflow planning, and execution-based repair. 
Starting from these intents, WorkFactory constructs a Semantic Workflow Graph that explicitly represents information requirements, dependencies, and locations within the task environment.
The graph guides the instantiation of task environments, while an adversarial execution–refinement loop exposes missing or inconsistent information and steers targeted repairs that preserve the original intent. 
Using WorkFactory, we construct 112,686 tasks spanning 612 occupations and train the WorkPilot family at 4B, 27B, and 35B-A3B scales on the resulting teacher trajectories.
All three models improve over their respective base models across four professional-work benchmarks, with WorkPilot-35B-A3B rising from 879 to 1565 Elo on GDPval-AA v2.1. 
Scaling experiments with up to 110k tasks further show that larger models benefit more from additional task data but exhibit diminishing returns at smaller data scales, while WorkPilot's long-horizon execution evolves from verbose trajectories toward more efficient task completion.
Controlled component studies quantify the negative effects of task-environment defects on downstream performance and show how semantic workflow planning and execution-based repair yield downstream gains through improved task-environment quality.
The subset of data are available at \url{https://github.com/IQuestLab/WorkFactory}.
\end{abstract}

\ifdefined\XeTeXversion
  \microtypesetup{tracking=false}
\fi

\begin{document}

\maketitle

\let\oldthefootnote\thefootnote

\begin{figure}[t]
  \centering
  \includegraphics[width=0.78\linewidth]{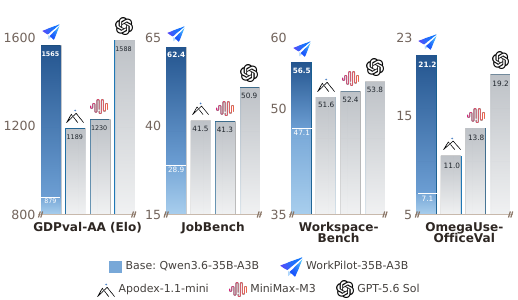}
  \caption{WorkPilot-35B-A3B consistently outperforms its Qwen3.6-35B-A3B base model across four professional-work benchmarks, with representative models shown for context.}
  \label{fig:benchmark-overview}
\end{figure}

\section{Introduction}
Language agents interleave reasoning and actions to interact with external environments~\citep{react}.
Working agents extend this paradigm to professional work in digital environments, executing workflows, coordinating tools, and producing deliverables across diverse occupations~\citep{gdpval,jobbench}.
This shift creates opportunities for agents to improve productivity and generate real economic value~\citep{mazeika2025remotelaborindexmeasuring}.
A common approach to agent post-training is supervised fine-tuning on expert demonstrations of task execution~\citep{song2026agent,deng2026fortsearchersynthesizingshortcutresistantsearch}.
Scaling this approach to working agents requires diverse professional intents and task environments that provide the information needed for teacher models to carry out professional workflows to completion.

Recent frameworks automate professional task construction by retrieving, reusing, adapting, or synthesizing task-relevant materials~\citep{syntheticcomputers}, and then collect teacher trajectories for supervision~\citep{nexforge}.
The value of scaling task construction, however, depends on whether the resulting environments provide sufficient and consistent information for teachers to demonstrate the target occupational competencies.
Establishing this support is challenging because the information required for an intended output may be distributed across multiple artifacts and connected through intermediate calculations and judgments.
For example, a supplier-selection task may require converting forecast demand into shipment quantities to determine the applicable vendor price tiers.
Although the demand forecasts and supporting logistics documents may each appear complete, a synthesized commercial summary may assign a discount inconsistent with the shipment quantities implied by these materials.
Detecting this discrepancy requires tracing the derivation from product demand to discount eligibility across files.
Task construction must therefore ensure that the supplied premises jointly support the intended analysis, while leaving intermediate calculations and the final recommendation to the teacher.
Otherwise, teachers may introduce unsupported assumptions or bypass essential analysis, weakening the resulting supervision.

In this paper, we introduce WorkFactory, a scalable framework for constructing executable professional tasks that support complete expert demonstrations. 
It organizes intent, workflow, and resources into an integrated task-construction process that provides the information needed for task completion while preserving the substantive work teachers must perform.
\textbf{(i) Diverse intent.} To cover common work demands as well as specialized roles and long-tail activities, we introduce \emph{hybrid intent sourcing}, combining occupational hierarchies, documented skills, and real professional deliverables to construct approximately 120K professional intents spanning 612 occupations.
\textbf{(ii) Grounded workflow.} We introduce a \emph{Semantic Workflow Graph} that jointly represents within-file dependencies and cross-file derivation chains, guiding synthesis at the level of information units rather than files alone.
Staged quality checks and an \emph{adversarial execution--refinement loop} test whether these dependencies survive environment construction.
Each repair is reviewed against the original intent to resolve support failures without eliminating essential work or leaking derived results.
\textbf{(iii) Flexible resources.} WorkFactory combines retrieved real-world artifacts with synthesized supporting materials, extending task construction beyond the work supported by existing files alone.
Resource acquisition proceeds on demand throughout construction, so additional materials can be retrieved or synthesized as specific information needs arise.
This hybrid construction yields 112,686 executable professional tasks.

Training Qwen3.5-4B~\citep{qwen3.5}, Qwen3.6-27B~\citep{qwen3.6-27b}, and Qwen3.6-35B-A3B~\citep{qwen36} on the resulting teacher trajectories yields the \emph{WorkPilot} family.
All three WorkPilot models substantially improve over their respective base models across four professional-work benchmarks, including GDPval-AA~\citep{gdpval,gdpvalaa} and JobBench~\citep{jobbench}, demonstrating the training utility of WorkFactory data across the tested model scales.
Figure~\ref{fig:benchmark-overview} illustrates the gains of WorkPilot-35B-A3B as a representative example.
Notably, WorkPilot-27B reaches a locally estimated 1612 Elo on GDPval-AA v2.1, placing it within the range of the compared frontier models.

By scaling professional task synthesis to over 110K tasks, WorkFactory enables controlled, in-depth analyses that yield practical guidance for constructing effective professional supervision.
\textbf{(i) Hybrid supervision scales, but its returns depend on the model.} Real and synthesized materials can jointly support effective training, providing a viable route to expanding professional supervision.
Model-dependent saturation patterns motivate tailoring data budgets to the learner, while continued gains alongside trajectories that first expand and then contract show that execution length alone is insufficient to assess learning progress.
\textbf{(ii) Semantic planning and execution-based repair address complementary failure modes.} Fine-grained planning improves derivability, while execution-based repair reduces latent inconsistencies and leakage in the realized environment.
Their combined 81-Elo gain supports pairing explicit dependency planning with execution feedback to prevent construction defects from propagating into teacher demonstrations.
\textbf{(iii) Teacher choice shapes downstream learning.} Given the same task environments, the stronger teacher yields an additional 235 Elo, showing that demonstration quality offers gains beyond task coverage.
Teacher-specific execution patterns are also reflected in students, motivating teacher selection for both capability and the behaviors transferred through supervision.

Our main contributions are as follows:
\begin{itemize}

\item \textbf{Scalable professional task construction.}
We introduce WorkFactory, a framework that turns diverse professional intents into executable task environments through hybrid intent sourcing, semantic workflow planning, and dynamic retrieval and synthesis of supporting materials.
A semantic workflow graph makes information dependencies explicit, while adversarial execution--refinement identify and repair environment defects, preserving the substantive calculations and judgments required of the teacher.

\item \textbf{Empirical insights into scalable and effective supervision.}
Experiments with up to 110K tasks demonstrate the supervision value of hybrid task environments, revealing model-dependent scaling gains and diminishing returns alongside an evolution toward more efficient long-horizon execution.
Controlled component studies connect improvements in environment quality to downstream gains from semantic planning and execution-based repair.
Matched-task teacher comparisons further show how teacher choice shapes both student performance and execution behavior.

\item \textbf{A large-scale professional task corpus and the WorkPilot family.}
We construct 112,686 tasks spanning 612 occupations and train WorkPilot models at 4B, 27B, and 35B-A3B scales.
All three models improve over their respective base models across four professional-work benchmarks, with WorkPilot-27B reaching a locally estimated 1612 Elo on GDPval-AA v2.1.
We release a subset of the task data to support further research.

\end{itemize}

\section{Related Work}
\label{sec:related_work}

\paragraph{Agent Task and Environment Synthesis.}
Synthetic task and environment construction has become an important approach to scaling agent training.
Software engineering provides a favorable setting, where repositories offer realistic contexts and executable tests provide concrete supervision; representative systems include SWE-Gym~\citep{swegym}, SWE-smith~\citep{swesmith}, and R2E-Gym~\citep{r2egym}.
Beyond coding, recent work synthesizes longer-horizon, multi-tool tasks through compositional expansion~\citep{agentsynth,taskcraft}, environment exploration and execution feedback~\citep{autoplay,dive}, and stateful interaction environments~\citep{agentworld}.
These advances motivate synthesis for working agents, whose tasks require coherent workspaces that integrate heterogeneous information and domain-specific deliverables.

\paragraph{Professional Task Construction for Working Agents.}
Table~\ref{tab:working-agent-pipelines} highlights differences in how professional task construction balances coverage, information support, and repair safeguards.
The compared pipelines derive intents primarily from one seed family: capability requirements in NexForge~\citep{nexforge}, personas in Synthetic Computers~\citep{syntheticcomputers}, occupational knowledge in GraphForge~\citep{GraphForge} and WorkGenesis~\citep{WorkGenesis}, or expert workflows in WorkForge~\citep{WorkForge}.
WorkFactory combines complementary seed families, extending coverage beyond the work elicited by any single route.
Workflow granularity exposes a further distinction: file-level plans in NexForge and Synthetic Computers organize materials, whereas information-level representations can specify the premises needed for particular derivations.
GraphForge and WorkForge exploit this finer granularity to ground tasks in existing real-file workspaces; WorkFactory uses it to guide the construction of both retrieved and synthesized materials.
This combines fine-grained information support with the flexibility of hybrid resources, allowing intended workflows to extend beyond what a coherent collection of existing files alone can support.
Repair safeguards distinguish the approaches further: GraphForge revises task statements and rubrics using execution feedback, while WorkGenesis repairs scenarios through rubric-level failure attribution.
WorkFactory additionally checks each repair against the original intent and plan, addressing the risk that an executable revision removes the professional work intended for supervision.
Its 110K-task corpus demonstrates that these construction and repair constraints can be maintained at scale.

\begin{table}[t]
  \centering
  \caption{Comparison of task-construction pipelines for working agents.}
  \label{tab:working-agent-pipelines}
  \footnotesize
  \setlength{\tabcolsep}{3pt}
  \renewcommand{\arraystretch}{1.22}
  \begin{tabularx}{\linewidth}{@{}>{\raggedright\arraybackslash}p{0.23\linewidth}cc>{\centering\arraybackslash}Xcc@{}}
    \toprule
    \textbf{Method} & \textbf{Intent} & \textbf{Workflow} & \textbf{Resources} & \makecell{\textbf{Work-preserving}\\\textbf{repair}} & \textbf{Scale} \\
    \midrule
    NexForge~\citep{nexforge} & Single & File-level & Real \& Synthetic & \xmark & 22K \\
    Synthetic Computers~\citep{syntheticcomputers} & Single & File-level & Real \& Synthetic & \xmark & 1K \\
    GraphForge~\citep{GraphForge} & Single & Information-level & Real & \xmark & 2.2K \\
    WorkGenesis~\citep{WorkGenesis} & Single & Not explicit & Real \& Synthetic & \xmark & 20K \\
    WorkForge~\citep{WorkForge} & Single & Information-level & Real & \xmark & 16.7K \\
    \midrule
    \rowcolor{iquestblue!6}
    \textbf{WorkFactory (Ours)} & \textbf{Hybrid} & \textbf{Information-level} & \textbf{Real \& Synthetic} & \cmark & \textbf{110K} \\
    \bottomrule
  \end{tabularx}
  \vspace{3pt}
  \begin{minipage}{\linewidth}
    \scriptsize
    \textit{Intent} distinguishes a single seed-source family from multiple complementary families (Hybrid).
    \textit{Workflow} indicates representation granularity: \textit{File-level} captures relations among files, whereas \textit{Information-level} represents facts, fields, or computational premises and their relations within and across files; \textit{Not explicit} means no separate dependency representation is specified.
    \textit{Resources} identifies whether input materials are real, synthetic, or both.
    \textit{Work-preserving repair} independently checks revisions against the original intent and plan (\xmark: not reported).
    \textit{Scale} is reported corpus size in thousands (K).
  \end{minipage}
\end{table}

\section{WorkFactory}
\label{sec:method}

\subsection{Overview}
\label{sec:overview}
\begin{figure}[t]
\centering
\includegraphics[width=0.86\linewidth]{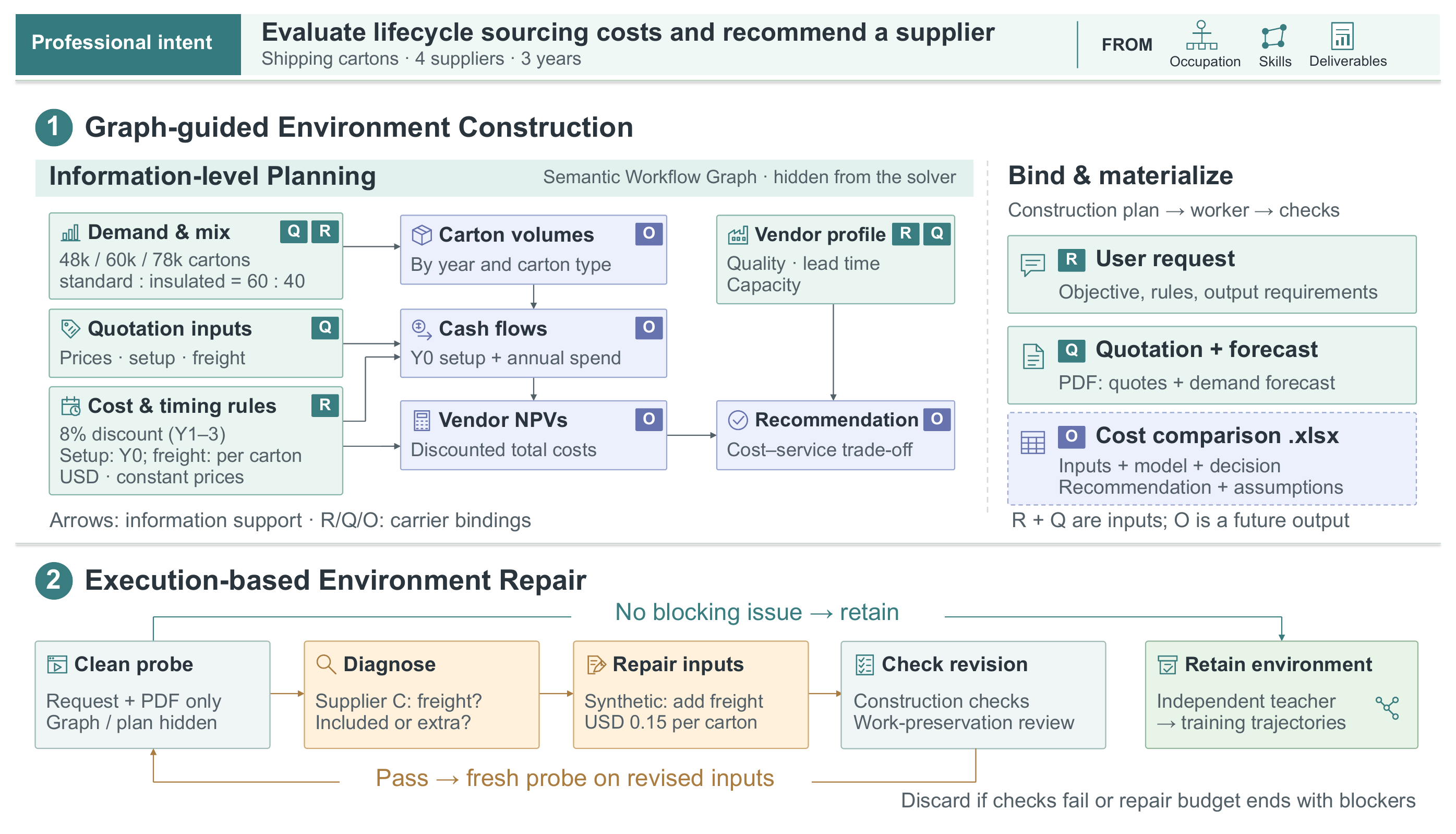}
\caption{
Overview of WorkFactory through a synthetic shipping-carton sourcing example.
\textbf{(1)} A Semantic Workflow Graph guides environment construction while leaving derived outputs to the solver.
\textbf{(2)} Clean probe executions expose blocking issues and guide checked repairs before retained environments are used for independent teacher trajectory collection.
Appendix~\ref{app:supplier-example} provides the complete end-to-end walkthrough.
}
\label{fig:workflow}
\end{figure}

WorkFactory constructs executable professional tasks for training working agents.
Hybrid intent sourcing builds a diverse intent library from occupational knowledge, documented skills, and real professional deliverables (\S\ref{sec:intents}).
For each intent, WorkFactory plans backward from the intended output using a Semantic Workflow Graph, which specifies information dependencies and their locations within and across files.
Guided by this graph, supporting materials are dynamically retrieved or synthesized to provide the required premises while leaving substantive derivations to the solver (\S\ref{sec:construction}).
Construction checks assess whether the resulting environment conforms to the plan, and an adversarial execution--refinement loop identifies and repairs blocking support failures while preserving the intended work (\S\ref{sec:refinement}).
Retained environments are assigned to an independent teacher for trajectory collection (\S\ref{sec:trajectories}).
Figure~\ref{fig:workflow} illustrates the process through a supplier-selection example.

\subsection{Hybrid Intent Sourcing}
\label{sec:intents}

Professional intents define the work to be performed before task-specific materials are constructed.
WorkFactory adopts \emph{hybrid intent sourcing} to capture common workplace demands while extending coverage to specialized and long-tail activities across 612 occupations.
It draws on three complementary sources: (1) \emph{occupational hierarchies}, which ground the work in roles and situations; (2) \emph{documented skills}, which identify operations that can be composed into workflows; and (3) \emph{real professional deliverables}, which provide cues about intended outputs and audiences.
These sources support distinct synthesis routes: expanding occupations into roles situated in professional contexts, composing skills through compatible input--output dependencies, and inferring work objectives from deliverables through instruction backtranslation~\citep{instructionbacktranslation}.
Candidate intents undergo LLM review for professional appropriateness, internal coherence, and feasibility as real workplace activities.
Accepted intents enter the shared library $\mathcal L$ to guide subsequent task construction.
Appendix~\ref{app:intent-synthesis} details the three synthesis routes.

\begin{figure}[t]
  \centering
  \includegraphics[width=0.90\linewidth]{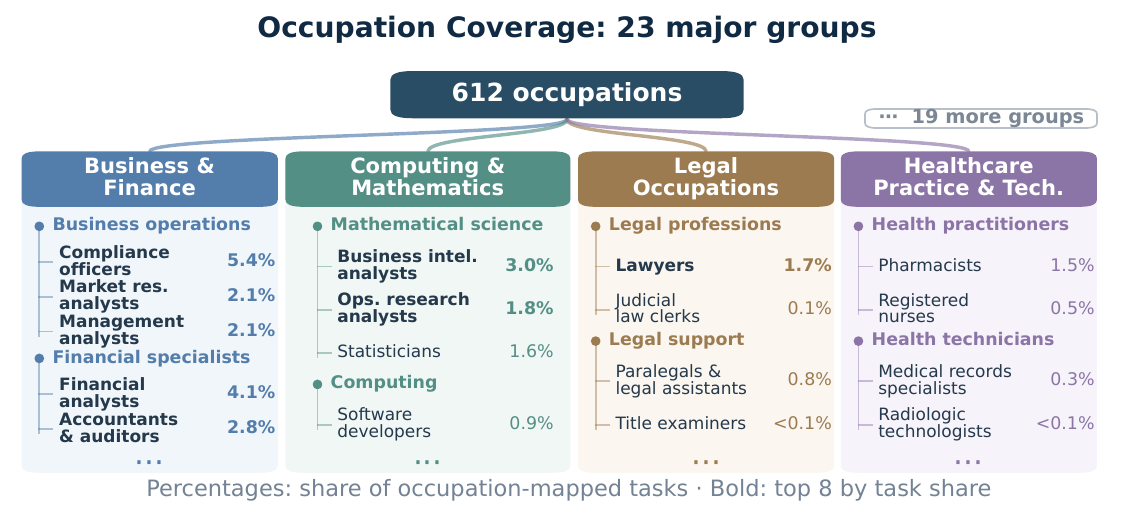}
  \caption{Selected high-coverage occupations from four of the 23 major occupational groups represented in the WorkFactory corpus.}
  \label{fig:occupation-coverage}
\end{figure}

\subsection{Graph-guided Environment Construction}
\label{sec:construction}

Translating a professional intent into executable work requires specifying the information needed for each deliverable and where that information should be available.
WorkFactory captures this structure through a \emph{Semantic Workflow Graph}, which traces intended outputs back to their supporting premises while leaving substantive calculations and judgments to the solver.
Beyond artifact-level dependencies, the graph represents semantic relations among information units within and across files.
It specifies the support structure of the intended work rather than a fixed sequence of solver actions.
Formally,
\begin{equation}
  \mathcal G_{\mathrm{info}} =
  (\mathcal V_{\mathrm{info}}, \mathcal V_{\mathrm{carrier}},
   \mathcal E_{\mathrm{info}}, \mathcal E_{\mathrm{loc}}),
  \label{eq:information-support-graph}
\end{equation}
where $\mathcal V_{\mathrm{info}}$ contains information units, $\mathcal V_{\mathrm{carrier}}$ identifies their carriers, $\mathcal E_{\mathrm{info}}$ represents semantic support relations among information units, and $\mathcal E_{\mathrm{loc}}$ binds those units to locations.

\paragraph{Planning information support.}
For each intent in $\mathcal L$, the planner starts from the intended output and works backward.
It identifies the calculations and judgments behind that output, then traces each of them to the information it requires.
The resulting information units form $\mathcal V_{\mathrm{info}}$, while $\mathcal E_{\mathrm{info}}$ records association, transformation, evidence, and constraint relations, including cases where several sources jointly support one result.
In the supplier-selection example in Figure~\ref{fig:workflow}, vendor NPVs rely on demand and pricing information distributed across the request and reference files; together with the vendor profiles, they support the final recommendation.
Each node records its content, entity and scope, and whether it is supplied, externally acquired, solver-derived, or internal to construction.
The source data and commercial terms are therefore made available, while the downstream calculations and recommendation remain unspecified for the solver to complete.

\paragraph{Grounding information in carriers.}
Location bindings in $\mathcal E_{\mathrm{loc}}$ map information to carriers in $\mathcal V_{\mathrm{carrier}}$: the user request, reference artifacts, external resources, future deliverables, or internal construction requirements.
They specify where information is supplied, acquired, or produced, and whether it is accessible to the solver.
Internal requirements remain hidden; any task-specific premise or obligation needed for completion must be expressed in or obtainable through solver-visible inputs.
In the supplier-selection example, input bindings locate demand forecasts and supplier quotations in a reference PDF and cost-timing rules in the request.
Output bindings associate the requested vendor NPV analysis and sourcing recommendation with a future cost-comparison workbook.
Appendix~\ref{app:information-support} provides detailed definitions.

\paragraph{Materializing the environment.}
The planner translates the graph into an artifact-level construction plan $\mathcal P=(u,\mathcal A,\mathcal D)$, comprising the concrete user request $u$, artifact specifications $\mathcal A$, and acyclic construction dependencies $\mathcal D$.
Guided by this plan, an environment worker dynamically retrieves suitable real-world artifacts for reuse or adaptation and synthesizes complementary materials when available files lack the required context or premises.
Resource acquisition thus follows the intended work rather than constraining tasks to a preassembled file collection.
Carrier bindings specify where information should appear, while semantic relations guide consistency across the constructed materials.
Programmatic checks verify file counts and types, and LLM review assesses content coverage and the coherence of the request with its reference files.
Failed checks trigger bounded construction retries; unresolved tasks are discarded.
Appendix~\ref{app:artifact-checks} details the procedure and retry budget.

\subsection{Execution-based Environment Repair}
\label{sec:refinement}
Execution-based repair implements an adversarial execution--refinement loop that challenges the constructed environment through independent task execution.
A probe diagnoses support failures from solver-visible inputs, and a refiner uses these findings to revise the environment.
Construction checks and a separate work-preservation review constrain each revision, preventing repairs from bypassing essential operations or supplying the solution.
The revised environment then undergoes a fresh probe execution.

\paragraph{Diagnosing the realized environment.}
A probe solver attempts the task using only the solver-visible request and resources, without access to the graph, construction plan, or prior execution outputs and diagnoses.
It grounds reported environmental issues in inspected sources and tool observations, identifies the affected operations and deliverables, and distinguishes them from solver mistakes and legitimate open choices.
Only issues judged to block task completion trigger repair; tasks without such blockers proceed to trajectory collection.
Appendix~\ref{app:defect-taxonomy} details the reporting criteria.

\paragraph{Repairing and revalidating the environment.}
A refiner uses the diagnoses to edit the request and existing reference files while preserving the files' number, types, and dependency structure; the Semantic Workflow Graph remains the initial construction specification.
For example, adding a missing USD~0.15-per-carton freight term for Supplier C restores the support needed for an NPV comparison without performing the comparison for the solver.
Each revision must pass construction checks and an LLM-based work-preservation review against the original intent and initial plan, assessing whether the core objective, essential operations, and substantive deliverable requirements remain intact.
Revised tasks that fail either check are discarded; those that pass undergo a fresh probe execution.
The loop stops when no blocker remains or the repair budget is exhausted, with unresolved tasks discarded.
Retained tasks form the candidate pool $\mathcal T$ for trajectory collection.
Appendix~\ref{app:repair-constraints} specifies the repair constraints and budget, and Appendix~\ref{app:implementation} provides model and execution settings.

\begin{figure}[t]
\centering
\begin{minipage}[b]{0.36\linewidth}
  \vspace{0pt}
  \centering
  \includegraphics[width=\linewidth]{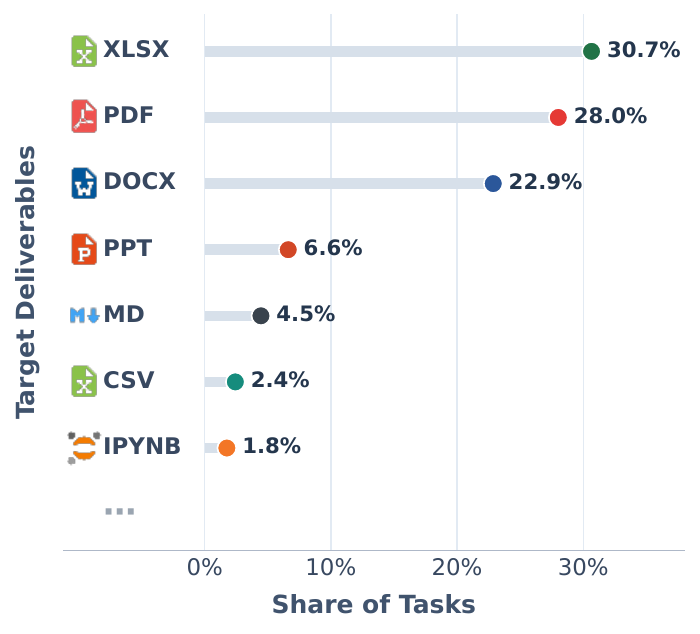}
  \par\vspace{4pt}
  {\small\textbf{(a) Deliverable distribution}\par}
\end{minipage}
\hfill
\begin{minipage}[b]{0.61\linewidth}
  \vspace{0pt}
  \centering
  \scriptsize
  \begin{tabular*}{\linewidth}{@{\extracolsep{\fill}}lr@{}}
    \toprule
    \textbf{Corpus statistic} & \textbf{Value} \\
    \midrule

    \rowcolor{blue!15} \multicolumn{2}{@{}l}{\textit{Tasks}} \\
    \# Tasks / Environments
        & 112,686 \\
    \# Occupations
        & 612 \\
    \# Deliverable File Extensions
        & 53$^{\dagger}$ \\

    \addlinespace[2pt]
    \rowcolor{blue!15} \multicolumn{2}{@{}l}{\textit{Workspaces}} \\
    Avg. Provided Reference Files / Task
        & 4.15$^{\ddagger}$ \\  
    Avg. Reference File Size
        & 1.56 MiB \\
    \# Reference File Extensions
        & 317$^{\dagger}$ \\

    \addlinespace[2pt]
    \rowcolor{blue!15} \multicolumn{2}{@{}l}{\textit{Teacher Trajectories}} \\
    Avg. Assistant Turns / Trajectory
        & 57.36 \\
    Avg. Tool Calls / Trajectory
        & 62.95 \\
    Avg. Total Tokens / Trajectory
        & 136.18K \\
    \bottomrule
  \end{tabular*}
  \par\vspace{3pt}
  \raggedright
  $^{\dagger}$File extensions occurring at least 10 times.
  $^{\ddagger}$Some tasks provide no reference files and require web search.
  \par\vspace{4pt}
  \centering
{\small\textbf{(b) Corpus statistics}\par}
\end{minipage}
\caption{Overview of the WorkFactory training corpus: (a) distribution of target deliverable file types; (b) aggregate statistics across tasks, workspaces, and teacher trajectories.}
\label{fig:corpus-overview}
\end{figure}

\subsection{Task Corpus and Training Data}
\label{sec:trajectories}

\paragraph{Teacher trajectory collection and supervision.}
For every task in $\mathcal T$, GLM-5.3~\citep{glm53} independently attempts the task in a clean sandbox with the same task-information access boundary as the probe in \S\ref{sec:refinement}.
The teacher receives the current request and accessible resources, without the Semantic Workflow Graph, construction plan, or prior probe outputs and diagnoses.
The resulting interaction trace and produced deliverables form an initial teacher rollout.
We retain rollouts that pass lightweight filters screening for repeated reasoning, repeated tool calls, and missing or insufficiently substantive deliverables.
For the 5\% of tasks with the longest initial rollouts, measured by total interaction turns, we additionally apply selective rejection sampling based on pairwise deliverable comparisons; all other tasks retain their filtered initial rollouts.
Each retained rollout is then converted into one or more training sequences for masked autoregressive supervision.\footnote{Context compaction and sub-agent strategy may split a rollout into multiple training sequences.}
Appendix~\ref{app:implementation} details the execution settings, rejection-sampling procedure, and training configuration.

\paragraph{Final training corpus.}
The final corpus contains 112,686 tasks spanning 612 fine-grained occupations and 23 major occupational groups; Figure~\ref{fig:occupation-coverage} shows the occupations with the highest task shares.
% For teacher execution, GLM-5.3 completes 94,077 of these tasks using the Stirrup harness~\citep{stirrup} and the remaining 18,609 using the Claude Code harness~\citep{claudecode}; Figure~\ref{fig:distribution-of-assistant-turns} shows the distribution of assistant turns per resulting trajectory.
We collect teacher trajectories using the Stirrup~\citep{stirrup} and the Claude Code harness~\citep{claudecode}.
Figure~\ref{fig:corpus-overview} shows the distribution of target deliverable file types alongside aggregate corpus statistics.
We screen the final corpus against the evaluation tasks using request-overlap checks, manual review of flagged pairs, and file hashes; these checks identify no evaluation-task reuse or byte-identical files shared with the training corpus.
Appendix~\ref{app:dataset-details} provides additional details,
while Appendix~\ref{app:synthesis-cost} reports construction-time token usage.

\section{Experiments}
\label{sec:exp}

\subsection{Experimental Setup}
\label{sec:setup}

\paragraph{Baselines.}
We compare the three WorkPilot models with their corresponding base models: Qwen3.5-4B~\citep{qwen3.5}, Qwen3.6-27B~\citep{qwen3.6-27b}, and Qwen3.6-35B-A3B~\citep{qwen36}.
% We evaluate WorkPilot-4B, WorkPilot-27B, and WorkPilot-35B-A3B from the \textbf{WorkPilot} family against their corresponding base models, Qwen3.5-4B~\citep{qwen3.5}, Qwen3.6-27B~\citep{qwen3.6-27b}, and Qwen3.6-35B-A3B~\citep{qwen36}.
These matched base-model comparisons serve as the primary controls for measuring the effect of WorkFactory supervision across model scales.
For broader system-level context, we also include three groups of representative models:
\textbf{frontier models}, including GLM-5.3~\citep{glm53}, DeepSeek-V4.1-Flash~\citep{deepseekai2026deepseekv41flashpushinglimitskv}, Qwen3.8-Max~\citep{qwen38}, GPT-5.6 Sol~\citep{openai2026gpt56}, and Kimi-K3~\citep{kimi-k3};
\textbf{open-weight models ($\mathord{\geq}300$B)}, including DeepSeek-V4-Pro-0813~\citep{deepseekv4}, DeepSeek-V4-Flash-0731~\citep{deepseekv4}, GLM-5.2~\citep{zai2026glm52}, and MiniMax-M3~\citep{minimaxm3};
and \textbf{open-weight models ($\mathord{\approx}30$B)}, including Qwen3.8-27B~\citep{qwen38}, Nex-N2.5-mini~\citep{nex25}, and Apodex-1.1-mini~\citep{an2026apodex11scalingagentic}.

\paragraph{Benchmarks.}
We evaluate the training utility of WorkFactory teacher trajectories on \textbf{GDPval}~\citep{gdpval}, \textbf{JobBench}~\citep{jobbench}, \textbf{Workspace-Bench}~\citep{workspacebench}\footnote{Following the official recommendation, we use Workspace-Bench-Lite for evaluation efficiency.}, and \textbf{OmegaUse-OfficeVal}~\citep{omegauseofficeval}.
For each locally evaluated model--task pair, we run three independent rollouts.
For GDPval, we follow the GDPval-AA protocol~\citep{gdpvalaa} with the Stirrup harness~\citep{stirrup}. 
We report official v2.1 Elo ratings where available; otherwise, using Gemini 3.1 Pro Preview~\citep{gemini31propreview} as the judge, we report local Bradley--Terry estimates~\citep{bradley1952rank} calibrated to the same v2.1 scale (see Appendix~\ref{app:two_anchor_elo}).
For the remaining benchmarks, we use the Claude Code harness~\citep{claudecode} with Claude Sonnet 4.6~\citep{anthropicsonnet46} as the judge when needed and report the official metrics.\footnote{Since the GLM-5.3 teacher supports text input only, we train and evaluate WorkPilot in text-only mode; the native vision and alignment components of its Qwen backbones remain unused.}

\paragraph{Training.}
All fine-tuned models, including those used in the controlled studies, use the shared supervised fine-tuning configuration reported in Appendix~\ref{app:implementation} (Table~\ref{tab:training-settings}).

\subsection{Effectiveness of WorkFactory Data}
\label{sec:main-results}

Table~\ref{tab:workfactory-benchmarks} presents the main results.
Fine-tuning on the same WorkFactory corpus improves the 4B, 27B, and 35B-A3B base models across all four professional-work benchmarks.
For example, WorkPilot-35B-A3B improves from 879 to 1565 on GDPval-AA, from 28.91 to 62.35 on JobBench, from 47.12 to 56.53 on Workspace-Bench, and from 7.14 to 21.18 on OmegaUse-OfficeVal.
On GDPval-AA, WorkPilot-27B reaches a locally estimated 1612 Elo; its point estimate is 12 Elo above DeepSeek-V4.1-Flash and 34 points below GLM-5.3, placing it within the range of the frontier models shown.
WorkPilot-4B also exceeds Apodex-1.1-mini on GDPval and OmegaUse-OfficeVal, demonstrating competitive performance at the 4B scale.
Together, the matched base-model gains demonstrate the training utility of WorkFactory teacher trajectories across the three model scales tested.

\begin{table*}[!ht]
  \centering
  \caption{Results on professional-task benchmarks. For GDPval-AA, $*$ marks an official v2.1 Elo rating; unmarked values are local two-anchor estimates, with task-bootstrap 95\% confidence intervals reported in Appendix Table~\ref{tab:local-elo-ci}.
  % Green parenthetical values show absolute score gains over the corresponding base model.
  }
  \label{tab:workfactory-benchmarks}
  \begingroup
  \small
  \setlength{\tabcolsep}{4pt}
  \renewcommand{\arraystretch}{1.17}
  \definecolor{benchmarkgroup}{gray}{0.94}
  \newcommand{\best}[1]{\textbf{#1}}
  \newcommand{\second}[1]{\underline{#1}}
  \newcommand{\official}[1]{#1\rlap{\textsuperscript{*}}}
  \newcommand{\gain}[1]{{\small\textcolor{green!55!black}{(+#1)}}}
  \begin{tabular}{@{}lcccc@{}}
    \toprule
    \multicolumn{1}{l}{\textbf{Model}} &
    \multicolumn{1}{c}{\textbf{GDPval-AA}} &
    \multicolumn{1}{c}{\textbf{JobBench}} &
    \multicolumn{1}{c}{\textbf{Workspace-Bench}} &
    \multicolumn{1}{c}{\textbf{OmegaUse-OfficeVal}} \\
    \midrule
    \rowcolor{benchmarkgroup}
    \multicolumn{5}{l}{\textit{Frontier Models}} \\
    GLM-5.3             & \official{1646} & 66.35 & 61.80 & 31.09 \\
    DeepSeek-V4.1-Flash & \official{1600} & 67.69 & 63.06 & 31.43 \\
    Qwen3.8-Max         & \official{1596} & 59.83 & 61.51 & 28.23 \\
    GPT-5.6 Sol         & \official{1588} & 50.91 & 53.77 & 19.25 \\
    Kimi-K3         & \official{1524} & 55.83 & 54.05 & 23.09 \\
    \midrule
    \rowcolor{benchmarkgroup}
    \multicolumn{5}{l}{\textit{Open-weight Models ($\geq 300$B)}} \\
    DeepSeek-V4-Pro-0813   & \official{1441} & 62.98 & 59.08 & 21.26 \\
    DeepSeek-V4-Flash-0731 & \official{1427} & 52.88 & 56.57 & 24.01 \\
    GLM-5.2                & \official{1358} & 48.08 & 54.00 & 20.72 \\
    MiniMax-M3             & \official{1230} & 41.32 & 52.43 & 13.81 \\
    \midrule
    \rowcolor{benchmarkgroup}
    \multicolumn{5}{l}{\textit{Open-weight Models ($\approx 30$B)}} \\
    Qwen3.8-27B      & \official{1409} & 62.12 & 59.26 & 23.47 \\
    Apodex-1.1-mini  & \official{1189} & 41.53 & 51.60 & 10.98 \\
    Nex-N2.5-mini    & 1127 & 20.92 & 44.81 & 10.91 \\
    \midrule
    \rowcolor{benchmarkgroup}
    \multicolumn{5}{l}{\textit{Our Models}} \\
    Qwen3.5-4B                  & 661 & 4.67 & 26.92 & 1.71 \\
    \addlinespace[1.5pt]
    \shortstack[l]{\hspace{1em}$+$ \textbf{WorkPilot-4B}\\[-2pt]
      \phantom{\gain{35.20}}}
      & \shortstack[c]{\textbf{1375}\\[-2pt]\gain{714}}
      & \shortstack[c]{\textbf{39.87}\\[-2pt]\gain{35.20}}
      & \shortstack[c]{\textbf{51.30}\\[-2pt]\gain{24.38}}
      & \shortstack[c]{\textbf{11.77}\\[-2pt]\gain{10.06}} \\
    Qwen3.6-35B-A3B             & \official{879} & 28.91 & 47.12 & 7.14 \\
    \addlinespace[1.5pt]
    \shortstack[l]{\hspace{1em}$+$ \textbf{WorkPilot-35B-A3B}\\[-2pt]
      \phantom{\gain{33.44}}}
      & \shortstack[c]{\textbf{1565}\\[-2pt]\gain{686}}
      & \shortstack[c]{\textbf{62.35}\\[-2pt]\gain{33.44}}
      & \shortstack[c]{\textbf{56.53}\\[-2pt]\gain{9.41}}
      & \shortstack[c]{\textbf{21.18}\\[-2pt]\gain{14.04}} \\
     Qwen3.6-27B                 & 975 & 37.81 & 49.41 & 10.18 \\
    \addlinespace[1.5pt]
    \shortstack[l]{\hspace{1em}$+$ \textbf{WorkPilot-27B}\\[-2pt]
      \phantom{\gain{25.24}}}
      & \shortstack[c]{\textbf{1612}\\[-2pt]\gain{637}}
      & \shortstack[c]{\textbf{63.05}\\[-2pt]\gain{25.24}}
      & \shortstack[c]{\textbf{59.18}\\[-2pt]\gain{9.77}}
      & \shortstack[c]{\textbf{26.61}\\[-2pt]\gain{16.43}} \\
    \bottomrule
  \end{tabular}
  \endgroup
\end{table*}

\subsection{Scaling Properties}
\label{sec:scaling}
We study how model performance and execution behavior scale with the number of distinct WorkFactory training tasks on GDPval under the GDPval-AA evaluation protocol.
Our primary sweep trains Qwen3.6-35B-A3B on nested subsets of 2K--110K tasks; we additionally evaluate Qwen3.6-27B and Qwen3.5-4B at 30K and 110K.

\begin{figure}[t]
  \centering
  \IfFileExists{figures/fig_scala_1.pdf}{%
    \includegraphics[width=0.90\linewidth]{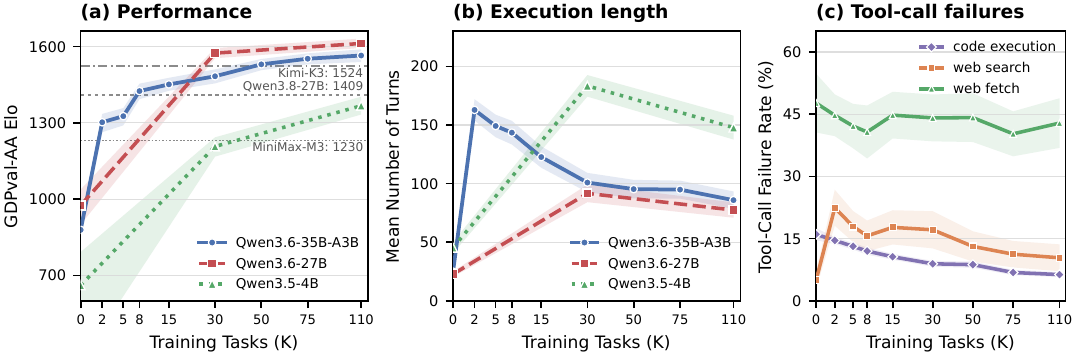}%
  }{%
    \fbox{\parbox[c][36mm][c]{0.96\linewidth}{\centering
      Placeholder for the data-scaling figure.}}%
  }
  \caption{
    Scaling on GDPval with the number of WorkFactory training tasks.
    \textbf{(a)} GDPval-AA Elo across model scales.
    \textbf{(b)} Task-level mean execution length.
    \textbf{(c)} Tool-call failure rates for Qwen3.6-35B-A3B.
    Shading denotes 95\% confidence intervals; the $x$-axis uses a square-root scale.
  }
  \label{fig:task-scaling}
\end{figure}

\paragraph{Performance improves with model-dependent diminishing returns.}
Figure~\ref{fig:task-scaling}(a) shows that GDPval-AA Elo increases with additional training data across all three models.
For Qwen3.6-35B-A3B, Elo rises from 879 to 1565: the first 5K tasks contribute 447 points, whereas scaling from 75K to 110K adds only 13.
By 30K tasks, the 35B-A3B and 27B models have captured 88\% and 94\% of their respective gains from the base model to the 110K-task checkpoint, showing different patterns of diminishing returns across the sampled model scales.
The 4B model also gains 546 Elo by 30K tasks, showing strong early data efficiency at the smallest tested scale.
A two-checkpoint comparison on JobBench shows the same direction: Qwen3.6-35B-A3B improves from 58.80 at 50K tasks to 62.35 at 110K.

\paragraph{Execution length first expands and then contracts.}
As shown in Figure~\ref{fig:task-scaling}(b), the mean execution length of Qwen3.6-35B-A3B first rises from 24.87 turns at baseline to 162.88 at 2K tasks, then falls to 85.90 at 110K while Elo continues to improve.
The fraction of its GDPval-AA trajectories reaching the 250-turn limit also falls from 21.9\% at 2K to 2.7\% at 110K.
Across the sampled checkpoints, the 27B and 4B models also show an increase from their bases to 30K tasks followed by a decrease at 110K; the 4B model uses longer trajectories than the other two models at both data scales.
Figure~\ref{fig:task-scaling}(c) further shows that, for Qwen3.6-35B-A3B, code-execution failures decline from 16.0\% at baseline to 6.3\% at 110K.
After rising to 22.4\% at 2K, web-search failures decrease to 10.4\% at 110K, whereas web-fetch failures remain at approximately 40\%--48\%.

Together, these observations show that performance and execution length evolve differently as training data increases: longer attempts emerge at small data scales, followed by shorter executions and lower failure rates for some tool types at larger scales.

\subsection{Component Analysis}
\label{sec:component-analysis}

We compare three 5K-task synthesis conditions: direct artifact-level planning, information-level planning using the Semantic Workflow Graph, and the same graph-based planning followed by execution-based repair.
The analysis asks whether explicitly planning and repairing task information improves both the conditions for teacher execution and the training utility of the resulting trajectories.
We assess task-environment quality using blocker- and correctness-level findings from a common probe solver (see \S\ref{sec:refinement}) and downstream utility by fine-tuning the same Qwen3.6-35B-A3B base model under the shared training configuration.
For each defect category, incidence is the fraction of tasks whose report contains at least one hard defect in that category; categories may overlap.
We use the initial probe report for the first two conditions and the final report for the repair condition, including tasks later discarded for unresolved blockers.
Table~\ref{tab:wf_repair_defects} summarizes both analyses.

\begin{table*}[!ht]
  \centering
    \caption{
    Component analysis across three 5K-task synthesis conditions.
    Defect entries report task counts (percentages), with categories allowed to overlap.
    Deltas compare adjacent conditions in percentage points for defects and GDPval-AA Elo points for downstream utility.
    }
\label{tab:wf_repair_defects}
  \small
  \setlength{\tabcolsep}{3pt}
  \renewcommand{\arraystretch}{1.04}
  \begin{tabular*}{\textwidth}{@{\extracolsep{\fill}}lccccc@{}}
    \toprule
    & \shortstack{\textbf{Artifact-level}\\\textbf{planning}}
    & \shortstack{\textbf{+ Semantic}\\\textbf{Workflow Graph}}
    & \shortstack{\textbf{$\Delta$ vs.}\\\textbf{left}}
    & \shortstack{\textbf{+ Execution-based}\\\textbf{repair}}
    & \shortstack{\textbf{$\Delta$ vs.}\\\textbf{left}} \\
    \midrule
    \multicolumn{6}{@{}l}{\textit{Probe-diagnosed hard-defect incidence}} \\
    Missing evidence         & 481 (9.62\%)     & 182 (3.64\%)     & \textcolor{green!55!black}{\ensuremath{-}5.98 pp}  & 101 (2.02\%) & \textcolor{green!55!black}{\ensuremath{-}1.62 pp} \\
    Cross-file inconsistency & 1,641 (32.82\%) & 1,287 (25.74\%) & \textcolor{green!55!black}{\ensuremath{-}7.08 pp}  & 242 (4.84\%) & \textcolor{green!55!black}{\ensuremath{-}20.90 pp} \\
    Broken derivation        & 1,243 (24.86\%) & 343 (6.86\%)     & \textcolor{green!55!black}{\ensuremath{-}18.00 pp} & 163 (3.26\%) & \textcolor{green!55!black}{\ensuremath{-}3.60 pp} \\
    Underspecified request   & 361 (7.22\%)     & 31 (0.62\%)      & \textcolor{green!55!black}{\ensuremath{-}6.60 pp}  & 14 (0.28\%) & \textcolor{green!55!black}{\ensuremath{-}0.34 pp} \\
    Leakage or shortcut      & 96 (1.92\%)      & 168 (3.36\%)     & \textcolor{red!70!black}{+1.44 pp}  & 42 (0.84\%) & \textcolor{green!55!black}{\ensuremath{-}2.52 pp} \\
    Execution feasibility    & 540 (10.80\%)    & 332 (6.64\%)     & \textcolor{green!55!black}{\ensuremath{-}4.16 pp}  & 24 (0.48\%) & \textcolor{green!55!black}{\ensuremath{-}6.16 pp} \\
    Self-inconsistency       & 164 (3.28\%)     & 223 (4.46\%)     & \textcolor{red!70!black}{+1.18 pp}  & 87 (1.74\%) & \textcolor{green!55!black}{\ensuremath{-}2.72 pp} \\
    Factual truth error      & 226 (4.52\%)     & 128 (2.56\%)     & \textcolor{green!55!black}{\ensuremath{-}1.96 pp}  & 65 (1.30\%) & \textcolor{green!55!black}{\ensuremath{-}1.26 pp} \\
    \midrule
    \multicolumn{6}{@{}l}{\textit{Post-training downstream utility}} \\
    GDPval-AA Elo            & 1245 & 1293 & \textcolor{green!55!black}{+48} & 1326 & \textcolor{green!55!black}{+33} \\
    \bottomrule
  \end{tabular*}
\end{table*}

\begin{figure}[t]
  \centering
  \includegraphics[width=0.60\linewidth]{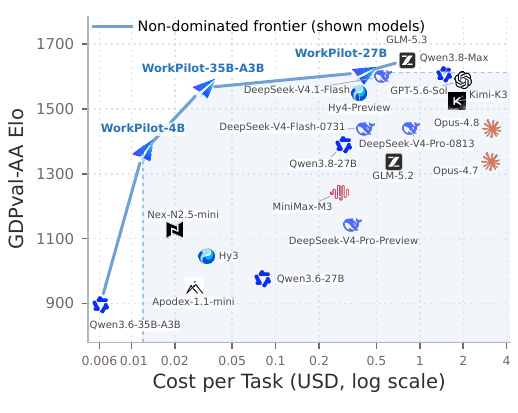}
  \caption{GDPval-AA Elo versus standardized estimates of model-inference cost per task. The line marks the non-dominated frontier among the models shown.}
  \label{fig:cost-pareto}
\end{figure}

\paragraph{Task-environment quality.}
Replacing direct artifact-level planning with information-level planning reduces probe-diagnosed hard-defect incidence in six of the eight categories.
The largest reduction is in broken derivations, from 24.86\% to 6.86\%, while underspecified requests fall from 7.22\% to 0.62\%.
These changes are consistent with the graph's role in making required information and its locations explicit.
However, the incidence of \emph{leakage or shortcuts} and \emph{self-inconsistency} increases by 1.44 and 1.18 percentage points, respectively.
We hypothesize that the former may arise when explicit dependencies lead the materializer to expose solver-derived information in the request, and the latter when denser constraints make it harder to maintain consistency across synthesized statements and values.
These regressions show that making information requirements explicit does not by itself ensure that materialization preserves them.
Adding execution-based repair reduces the reported incidence in all eight categories relative to information-level planning alone.
In particular, cross-file inconsistency falls from 25.74\% to 4.84\%, execution-feasibility defects from 6.64\% to 0.48\%, and leakage or shortcuts from 3.36\% to 0.84\%.
Taken together, these reductions show that execution-based repair complements information-level planning by addressing defects that remain in the materialized environments.
Appendix~\ref{app:repair-dynamics} provides additional repair statistics.

\paragraph{Downstream utility.}
We train the same model separately on teacher trajectories from each synthesis condition.
Information-level planning raises GDPval-AA Elo from 1245 to 1293, while adding execution-based repair further raises it to 1326.
The cumulative 81-point gain over direct artifact-level planning accompanies reductions in probe-diagnosed defects, linking improvements in task construction to the downstream utility of the collected teacher trajectories.

\subsection{Effect of Teacher Choice}
\label{sec:teacher-sensitivity}

We compare supervision from two teachers on the 30K-task subset from the scaling study (\S\ref{sec:scaling}).
The GLM-5.3 condition reuses the Qwen3.6-35B-A3B checkpoint trained on that subset; the DeepSeek condition regenerates trajectories for the same tasks with DeepSeek-V4-Flash-0731 and fine-tunes the same base model with the shared configuration.
The student trained on GLM-5.3 trajectories reaches 1483 Elo ($[1455,1510]$), whereas the student trained on DeepSeek-V4-Flash-0731 trajectories reaches 1248 Elo ($[1220,1274]$); both outperform the 879-Elo base model.
Both teacher conditions improve over the base model, supporting the utility of WorkFactory tasks beyond a single teacher.
The difference between students also shows that trajectory-generation choices affect how much training benefit is obtained from the same task set.
Appendix~\ref{app:teacher-sensitivity} analyzes the corresponding teacher and student behaviors.

\subsection{Cost--Performance Trade-off}
\label{sec:cost-performance}

We estimate model-inference cost per GDPval-AA task from recorded input and output tokens using a common accounting formula and explicit mappings to provider prices.
The estimates use a shared cache-hit assumption and explicit price proxies where needed, and cover task-solving model inference only.
Under this standardized accounting, all three WorkPilot models lie on the non-dominated frontier among the models shown in Figure~\ref{fig:cost-pareto}; the estimated costs of WorkPilot-4B, WorkPilot-35B-A3B, and WorkPilot-27B are approximately \$0.012, \$0.031, and \$0.396 per task, respectively.
Appendix~\ref{app:cost-performance} details the accounting method and price mappings.

\section{Conclusion}
\label{sec:conclu}
We introduced WorkFactory, a scalable framework that combines hybrid intent sourcing with dynamic retrieval and synthesis of supporting materials to construct diverse professional tasks.
A Semantic Workflow Graph specifies fine-grained information dependencies, while execution-based repair checks whether the constructed environment supports the intended work without removing its essential operations.
Using teacher trajectories from 112,686 tasks spanning 612 occupations, we trained the WorkPilot family, whose three models improve over their corresponding bases across four professional-work benchmarks.
Data-scaling experiments reveal diminishing performance returns and execution lengths that first expand and then contract.
Controlled component studies show that semantic planning and execution-based repair improve environment quality and downstream utility, while matched-task teacher comparisons highlight the importance of the demonstrations collected from the same tasks.
Together, these findings support coupling broad task coverage with validation throughout construction and careful selection of teacher supervision.

\section*{Limitations}

First, GLM-5.3, our teacher model, accepts only text input. Consequently,
although the Qwen backbones include native vision and alignment components, we
train and evaluate them only through a text-only task-solving interface. This
work therefore does not examine learning from visual evidence directly. Future
work will extend trajectory collection and student training to multimodal
professional tasks.
Second, our training study focuses on supervised fine-tuning over teacher
trajectories.
Future work will leverage high-quality professional tasks and their execution
feedback to train working agents with reinforcement learning.

\section*{Ethical Considerations}

WorkFactory trains agents to perform professional workflows and produce
realistic workplace artifacts. Such outputs may be incorrect, biased, or
misused in consequential settings, and should therefore receive appropriate
human review rather than replace qualified professionals. Dataset and model
release should also respect privacy, copyright, and licensing requirements.

% \clearpage
% \newpage

\bibliography{custom}
\newpage
\appendix
\flushbottom
% Appendix content. The main manuscript invokes \appendix.

\section{Professional Intent Synthesis}
\label{app:intent-synthesis}
\suppressfloats[t]

\paragraph{Occupation-driven synthesis.}
Starting from a public occupational hierarchy derived from the Standard
Occupational Classification~\citep{soc2018}, we expand broad occupations
into fine-grained roles. These roles seed a search agent that retrieves
professional context from the web as needed and generates work situations
and objectives grounded in role responsibilities, business settings, and
motivating events.

\paragraph{Skill-driven synthesis.}
We collect documented skills from a public skill directory\footnote{\url{https://skills.sh/}}
and analyze each skill's purpose, applicability conditions, prerequisites,
and operations to derive work situations and objectives. We combine skills
into candidate multi-step workflows when one skill's output can supply inputs
required by another and the operations jointly serve a coherent professional
objective. For example, quotation extraction supplies supplier cost inputs for
annual cash-flow modeling, whose results feed NPV comparison and sourcing
recommendation.

\paragraph{Deliverable-driven synthesis.}
Instruction backtranslation generates candidate instructions for existing
text~\citep{instructionbacktranslation}. Inspired by this, we use real
professional deliverables collected from the web as cues for plausible work
intents. We analyze their likely use, audience, content, and implied
production process to derive intents for producing, updating, reviewing, or
comparing such artifacts.

\paragraph{Intent review.}
An LLM reviewer evaluates whether each candidate intent is coherent and
feasible, professionally appropriate, and plausible as a real workplace activity.
Candidates that fail this review are excluded from the shared intent library.

\begin{table}[t]
  \centering
  \caption{Information-node roles.}
  \label{tab:info-node-roles}
  \small
  \setlength{\tabcolsep}{2pt}
  \renewcommand{\arraystretch}{1.08}
  \begin{tabularx}{0.99\linewidth}{@{}>{\raggedright\arraybackslash}p{0.25\linewidth}YY@{}}
    \toprule
    \textbf{Node role} & \textbf{Meaning} & \textbf{Example} \\
    \midrule
    Facts / evidence & Records and observations & Demand forecasts and supplier quotations \\
    Rules / definitions & Policies and conventions & Setup, freight, and discount rules \\
    Intermediate & Values to derive & Carton volumes, annual spend, and vendor NPVs \\
    Deliverable & Requested work product & Cost-comparison workbook and recommendation \\
    Requirements & Goals and constraints & Compare lifecycle cost and vendor service \\
    \bottomrule
  \end{tabularx}
\end{table}

\begin{table*}[htbp]
\centering
\caption{Information support for the synthetic shipping-carton sourcing example.
Inputs are supplied in the request and reference PDF; carton volumes, spending,
NPVs, and the recommendation remain solver work.}
\label{tab:procurement-support}
\small
\setlength{\tabcolsep}{4pt}
\renewcommand{\arraystretch}{1.12}
\begin{tabularx}{\linewidth}{@{}>{\raggedright\arraybackslash}p{0.28\linewidth}>{\raggedright\arraybackslash}p{0.24\linewidth}Y@{}}
\toprule
    \textbf{Information unit} &     \textbf{Carrier / status} &     \textbf{Support relation} \\
\midrule
Lifecycle-cost analysis and recommendation requirements & User request; input & Specify the analysis and workbook to produce. \\
Annual carton demand and standard/insulated mix & Quotation and forecast PDF; input & Determine annual carton volumes by type. \\
Supplier unit prices, setup fees, and freight terms & Quotation and forecast PDF; input & Supply vendor-specific cost inputs. \\
Discount and cost-timing rules & User request; input & Place setup fees in year 0 and discount years 1--3 at 8\%. \\
Vendor quality, lead time, and capacity & Request and quotation; input & Constrain and contextualize the sourcing decision. \\
Carton volumes, year-0 setup, annual spend, and vendor NPVs & Cost-comparison workbook; to be derived & Transform demand, quotations, and rules into comparable lifecycle costs. \\
Recommendation and assumptions & Cost-comparison workbook; to be produced & Combine vendor NPVs with the vendor profiles. \\
Construction requirements & Internal specification; hidden & Organize the request, reference PDF, and future workbook without supplying derived values. \\
\bottomrule
\end{tabularx}
\end{table*}

\section{Information-level Planning and Task Instantiation}
\label{app:information-support}
WorkFactory represents the information required by a task together with the
artifacts that provide or receive it. The representation distinguishes a
planned output or internal construction requirement from information already
available to the solver. Internal construction requirements serve only
as an intermediate specification during synthesis, connecting the intended
work to the request and supporting materials. They contain no instance-specific
facts and add no obligations unavailable to the solver.
All goals, constraints, and instance-specific information needed for completion
must be expressed in or obtainable through solver-accessible inputs.

\subsection{Information Nodes and Relations}
An information node records semantic content, entity and scope, and provision
status: supplied, externally acquired, solver-derived, or internal to
construction. It also records carrier bindings specifying where the content
is supplied, acquired, or produced. Solver-derived nodes describe quantities
or outputs to be derived and their supporting relations; their values remain
to be determined by the solver. Table~\ref{tab:info-node-roles} summarizes the principal roles.
A relation is $r=(S_r,v_r,\rho_r)$, where source nodes $S_r$ jointly support
target node $v_r$ under relation $\rho_r$. Relations capture reference and
consistency, entity association, scope and applicability, calculation,
evidence support, constraints, and requirement fulfillment. They may connect
content within one file, across files, or between the request, evidence,
deliverable, and internal construction requirements.
Table~\ref{tab:procurement-support} instantiates these nodes and relations for
the supplier-selection example.

\subsection{Carriers and Access}
Carrier nodes identify where information occurs: the user request, reference
artifacts, external resources, future deliverables, or internal construction requirements.
A location binding records the carrier, provision status, and access
conditions. Before materialization it specifies a planned location; after
construction it can resolve to a file, sheet, cell range, page, or paragraph.
The same information may occur in several carriers. A shared identity and
scope link these occurrences and guide the worker in maintaining consistency
across files. Access metadata distinguishes solver-visible content from
internal construction requirements. Task-specific premises needed by the solver
must be supplied in accessible materials or obtainable from external resources.
Future deliverables are specified through their required content and supporting
relations, leaving their production to the solver.

\subsection{Artifact Construction and Checks}
\label{app:artifact-checks}
The artifact-level construction plan $\mathcal P=(u,\mathcal A,\mathcal D)$
contains the concrete user request and deliverable requirements $u$, artifact
specifications $\mathcal A$, and acyclic construction dependencies $\mathcal D$.
Each artifact specification records its role, format, required content,
provenance, and materialization route: reuse, adaptation, or synthesis.
The environment worker receives both the Semantic Workflow Graph and this
plan as text in its planning context, and materializes all required artifacts
within a single agent session.
The worker dynamically retrieves suitable real artifacts for reuse or adaptation
and synthesizes components when usable sources lack required
information or the task requires a controlled condition such as a resolvable
conflict. Artifact construction dependencies specify when one file needs
content from another. The worker follows these dependencies as prerequisite
constraints in its prompt when deciding the construction order.
Other information relations remain constraints in the
shared planning context and may be mutual.
Programmatic checks compare planned and generated file counts and types.
LLM-based review assesses content coverage and the coherence of the user
request, including its references to all provided reference files.
During initial materialization, failed checks trigger up to three automatic
construction retries after the initial attempt, stopping when the checks pass;
tasks that still fail at the limit are discarded.
Passing the construction checks does not establish that every semantic dependency is
satisfied; execution feedback supplies an additional check on the realized
environment.

\begin{table*}[t]
  \centering
  \caption{Defect categories assigned retrospectively to issues in execution-based quality reports.}
  \label{tab:defect-taxonomy}
  \small
  \setlength{\tabcolsep}{5pt}
  \renewcommand{\arraystretch}{1.08}
  \begin{tabularx}{\textwidth}{@{}p{0.25\textwidth}Y@{}}
    \toprule
    \textbf{Category} & \textbf{Definition} \\
    \midrule
    Missing evidence & Required information is absent from all solver-accessible sources. \\
    Cross-file inconsistency & Independent sources disagree on entities, values, periods, units, or definitions. \\
    Broken derivation & A requested result cannot be derived from the supplied premises or rules. \\
    Underspecified request & The request omits a decision needed to determine a valid deliverable. \\
    Leakage or shortcut & Inputs reveal an answer or internal evaluation signal that should require work. \\
    Execution feasibility & A hard constraint, corrupt artifact, or unavailable tool makes completion impossible. \\
    Self-inconsistency & Statements or records within one artifact or request contradict each other. \\
    Factual truth error & A supplied fact, mapping, standard, or dated value conflicts with an identified authoritative source. \\
    \bottomrule
  \end{tabularx}
\end{table*}

\section{Execution-based Environment Repair}
\label{app:repair-details}

\subsection{Probe Reporting and Defect Taxonomy}
\label{app:defect-taxonomy}
Runtime quality assessment reports describe issues and assign severity labels.
A probe solver is informed that the task is synthetic and instructed to produce the best feasible deliverable.
It grounds each diagnosis in inspected sources and tool observations, identifies affected operations and deliverable requirements, and distinguishes environmental defects from solver mistakes and legitimate open choices.
Severity labels distinguish \textit{blocker} findings that prevent completion
of the minimum required deliverable, \textit{correctness} findings that identify
concrete task errors, \textit{efficiency} findings that suggest better
processes, and \textit{open} observations that record non-binding risks or
suggestions. Only blocker findings trigger repair, although a triggered repair
may also address correctness-level and other reported issues. Tasks with only
non-blocking findings skip repair; efficiency and open findings are advisory.
Non-blocking errors may therefore remain in retained tasks, and unclassified
findings do not enter the eight-category table.

For the component analysis, we retrospectively categorize the issues described in these reports by their content and supporting evidence.
The eight categories in Table~\ref{tab:defect-taxonomy} are therefore analysis labels, separate from the severity labels used to trigger repair.
In particular, the category \textit{Execution feasibility} describes execution-related obstacles; blocker-severity issues can also concern missing evidence, inconsistent materials, or other categories.
Blocker and correctness findings are treated as hard defects in the incidence
analysis. Task-level incidence counts a task once per category if its selected
report contains at least one finding labeled \textit{blocker} or
\textit{correctness} in that category; a task may appear in multiple categories.
Classification uses the report's stated evidence, not solver failure alone.
Table~\ref{tab:wf_repair_defects} uses the initial report (round 0) for the
artifact-level and information-level planning conditions and the final report for the
execution-based repair condition, including tasks subsequently discarded for
unresolved blockers.

\subsection{Work-Preservation and Repair Constraints}
\label{app:repair-constraints}
The original professional intent and the initial construction plan are the
references for reviewing whether a repaired task retains its
core objective, essential operations, and substantive deliverable requirements.
Minor changes to the work's scope or details are permitted when restoring
the conditions needed for completion; the initial plan is a reference for
the intended work, rather than a requirement to preserve every instance detail.
The refiner directly edits the request $u$ and the contents of existing
reference files specified by $\mathcal A$, without updating the Semantic Workflow Graph. Its prompt requires the number, dependency relations, and
types of reference files to remain unchanged.
For example, adding the missing Supplier C freight term of USD~0.15 per carton
preserves the lifecycle-sourcing objective, whereas dropping the vendor NPV
comparison or supplying a completed recommendation removes essential work.
This separate LLM-based review is applied after
each repair, rather than during initial materialization; tasks that require
no repair do not undergo it.

\begin{table*}[!t]
  \centering
  \caption{Repair depth and paired probe-execution statistics for the execution-based repair condition on the 5K-task analysis set.}
  \label{tab:repair-dynamics}
  \small
  \setlength{\tabcolsep}{5pt}
  \renewcommand{\arraystretch}{1.08}
  \begin{tabular*}{\textwidth}{@{\extracolsep{\fill}}lrr@{}}
    \toprule
    \multicolumn{3}{@{}l}{\textit{Repair depth}} \\
    \textbf{Rounds} & \textbf{Tasks} & \textbf{Share} \\
    \midrule
    0         & 3,479 & 69.58\% \\
    1         & 1,012 & 20.24\% \\
    2         &   385 &  7.70\% \\
    3 (limit) &   124 &  2.48\% \\
    \midrule
    \multicolumn{3}{@{}l}{\textit{Paired probe execution statistics}} \\
    \textbf{Metric} & \textbf{Before} & \textbf{After / change} \\
    \midrule
    Assistant turns          & 57.67     & 58.35 / +1.18\% \\
    Assistant output tokens  & 84,732.75 & 88,170.80 / +4.06\% \\
    Total trajectory tokens  & 136,638.74 & 142,363.28 / +4.19\% \\
    \bottomrule
  \end{tabular*}
\end{table*}

Each revision also undergoes the construction checks in
Appendix~\ref{app:artifact-checks}.
Revised tasks that fail construction or work-preservation checks are discarded.
Revisions that pass are re-executed by the probe in a clean sandbox without
access to earlier outputs or reports. Up to three repair rounds are allowed;
tasks with unresolved blockers after the final round are discarded.
The same material can play different roles under different intents: a
precomputed comparison may be a valid input to an audit but would remove
required work from a task asking the solver to perform that comparison.
Likewise, conflicting records can be legitimate inputs when resolving the
conflict is part of the intended work. Information repeated across carriers
should otherwise agree in identity, scope, and value. The representation
specifies supporting evidence and constraints without prescribing a unique
judgment or execution strategy.

\subsection{Repair Dynamics on the 5K Analysis Set}
\label{app:repair-dynamics}

The component analysis in \S\ref{sec:component-analysis} compares three
5K-task synthesis conditions: direct artifact-level planning,
information-level planning, and information-level planning followed by
execution-based repair. It evaluates pre-training environment quality using
hard defects identified by a common probe solver and downstream utility by
training the same student model under each condition. This subsection further
characterizes the repair condition by reporting repair depth and paired probe
execution statistics on the same 5K-task analysis set.
Table~\ref{tab:repair-dynamics} summarizes both views.

Tasks without a blocker-level finding skip refinement; 20.24\% undergo one
repair, 7.70\% undergo two, and 2.48\% reach the three-round limit. Of the
latter group, 24 tasks (0.48\% of the full set) remain blocked and are excluded
from training. After repair, mean reasoning tokens decrease by 1,738.58,
tool-call argument tokens increase by 5,265.34, and body-text tokens decrease
by 88.70, for a net increase of approximately 3,438 assistant-output tokens.
These length statistics describe a shift toward tool interaction.

\subsection{End-to-End Supplier-Selection Example}
\label{app:supplier-example}
The following synthetic instance traces the running example from planning to
training. The intent is to evaluate lifecycle sourcing costs and recommend a
supplier for shipping cartons across four anonymous candidate vendors and a
three-year program. During instantiation, the planner specifies annual demand
of 48K, 60K, and 78K cartons and a 60:40 standard-to-insulated mix. Supplier
quotations provide unit prices, setup fees, and freight terms, while visible
rules specify an 8\% discount rate for years 1--3, one-time setup costs in year
0, per-carton freight, USD-denominated constant prices, and no discounting of
the year-0 setup fees.

The graph marks the demand forecast, quotations, cost-timing rules, and vendor
profiles as supplied information. Transformation relations connect demand and
carton mix to annual carton volumes, then combine those volumes with each
vendor's quotation to derive year-0 setup, annual spend, and vendor NPVs. The
NPVs and vendor profiles jointly support a cost--service recommendation.
Carrier bindings place the objective and rules in the user request, the
quotations and demand forecast in a reference PDF, and the inputs, model,
decision, recommendation, and assumptions in a future cost-comparison workbook.
The carton volumes, spending, NPVs, and recommendation remain solver-derived.

In the realized environment, Supplier C's quotation does not state whether
freight is included or charged separately. A clean probe can diagnose that the
visible inputs do not support a comparable lifecycle-cost calculation. The
refiner adds a synthetic term specifying an additional freight charge of
USD~0.15 per carton to the quotation PDF, after which construction checks and
work-preservation review are repeated. A fresh probe then attempts the revised
task without seeing the earlier diagnosis. Neither initial construction nor
repair supplies the spending calculations, vendor NPVs, or final
recommendation.

Finally, an independent teacher completes the cost-comparison workbook across
all four vendors and produces a justified sourcing recommendation from the
solver-visible inputs. The resulting rollout is processed for training as described in
Appendix~\ref{app:implementation}.

\section{Corpus and Training Details}
\label{app:dataset-details}

\subsection{Corpus Statistics}
\begin{figure}[t]
  \centering
  \includegraphics[width=0.62\linewidth]{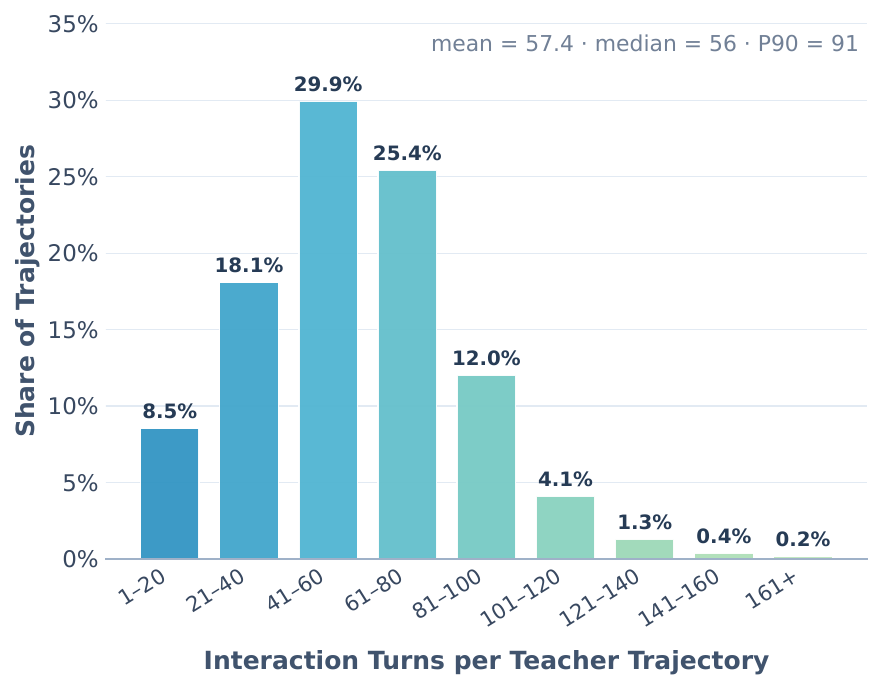}
  \caption{Distribution of assistant turns per teacher trajectory in the WorkFactory corpus.}
  \label{fig:distribution-of-assistant-turns}
\end{figure}

The final corpus represents 112,686 distinct tasks, each with at least one
training sequence. Before sequence construction, teacher rollouts undergo
lightweight validity filtering and, for selected long rollouts, rejection
sampling; overlength sequences are then removed during preprocessing. For teacher execution,
GLM-5.3 completes 94,077 of these tasks using GDPval Stirrup and the remaining
18,609 using Claude Code. The corpus spans 612
fine-grained occupations and contains 4.15 provided reference files per task
on average. Among file extensions occurring at least 10 times, it covers 317
reference-file extensions and 53 deliverable-file extensions.
Figure~\ref{fig:distribution-of-assistant-turns} shows that teacher trajectory
lengths are concentrated around 41--80 assistant turns, with a mean of 57.4,
a median of 56, and a 90th percentile of 91 turns.

\paragraph{Benchmark decontamination.}
Evaluation tasks are not exposed to WorkFactory during intent synthesis or
task-environment construction. As a post-hoc safeguard, we encode the user
request of every retained training task and every evaluation task with
Qwen3-Embedding-8B~\citep{qwen3embedding} and flag pairs with cosine
similarity of at least 0.85. Separately, we flag pairs when shared 10-grams
cover at least 20\% of the tokens in the shorter request. We also compare
hashes of the solver-visible files in the training and evaluation environments
to detect byte-identical artifacts. All request pairs flagged by either
text-overlap criterion are manually reviewed to distinguish reuse of an
evaluation task from benign similarity in domain, occupation, or deliverable
type. No reviewed pair constitutes task reuse, and no file hash is shared
between the training and evaluation environments.

\subsection{Construction Cost}
\label{app:synthesis-cost}
We measure construction cost in model-token usage.
Across the three primary environment-construction stages, recorded model
usage averages 11.656M tokens per attempted task: 11.448M input tokens and
0.208M output tokens.
Table~\ref{tab:synthesis-cost} reports the per-stage breakdown. Information-level planning is the largest
component at 4.737M tokens per task, followed by execution and repair at
3.644M and material synthesis at 3.275M. 

\begin{table}[!htbp]
  \centering
  \caption{Mean model-token usage per task by construction stage. Token counts are in millions.}
  \label{tab:synthesis-cost}
  \small
  \setlength{\tabcolsep}{3pt}
  \begin{tabular*}{\columnwidth}{@{\extracolsep{\fill}}lrrr@{}}
    \toprule
    \textbf{Synthesis stage} & \textbf{Input} & \textbf{Output} & \textbf{Total} \\
    \midrule
    Information-level planning & 4.658 & 0.079 & 4.737 \\
    Material synthesis          & 3.219 & 0.056 & 3.275 \\
    Execution and repair        & 3.571 & 0.073 & 3.644 \\
    \midrule
    Total                       & 11.448 & 0.208 & 11.656 \\
    \bottomrule
  \end{tabular*}
\end{table}

\subsection{Trajectory Generation and Training}
\label{app:implementation}
\paragraph{Model backends.}
All agents used for intent synthesis, information-level planning, environment
construction, and execution-based repair use
DeepSeek-V4-Flash-0731~\citep{deepseekv4}.
GLM-5.3~\citep{glm53} serves as the independent teacher model for trajectory
collection.

\paragraph{Training configuration.}
We initialize WorkPilot-4B, WorkPilot-27B, and WorkPilot-35B-A3B from
Qwen3.5-4B~\citep{qwen3.5}, Qwen3.6-27B~\citep{qwen3.6-27b}, and
Qwen3.6-35B-A3B~\citep{qwen36}, respectively. All three models are trained on
the selected GLM-5.3~\citep{glm53} task-solving trajectories using the same
configuration. Table~\ref{tab:training-settings} summarizes the probe,
teacher-generation, and supervised fine-tuning settings.
All three Qwen base checkpoints are distributed under the Apache License 2.0,
which permits modification and redistribution of derivative works subject to
its notice and attribution requirements. We release the corresponding
WorkPilot checkpoints with the applicable license and attribution notices
retained and clearly identify them as fine-tuned derivatives.

\begin{table}[t]
  \centering
  \caption{Probe execution, teacher trajectory generation, and shared WorkPilot training settings.}
  \label{tab:training-settings}
  \small
  \setlength{\tabcolsep}{4pt}
  \renewcommand{\arraystretch}{1.08}
  \begin{tabularx}{\linewidth}{@{}>{\raggedright\arraybackslash}p{0.43\linewidth}Y@{}}
    \toprule
    \textbf{Setting} & \textbf{Value} \\
    \midrule
    \multicolumn{2}{@{}l}{\textit{Probe execution}} \\
    Probe model & DeepSeek-V4-Flash-0731 \\
    Tool interface & Claude Code tools \\
    Context window & 1M tokens \\
    Context compaction & Disabled \\
    \midrule
    \multicolumn{2}{@{}l}{\textit{Teacher trajectory generation}} \\
    Teacher model & GLM-5.3 \\
    Execution harness (tasks) & GDPval-AA Stirrup (94,077); Claude Code (18,609) \\
    Tool interface & GDPval-AA tools (Stirrup); native tools (Claude Code) \\
    Context window & 256K tokens \\
    Context compaction & GDPval-AA mechanism (Stirrup); threshold 0.7 (Claude Code) \\
    Sampling & Temperature 1.0; top-$p$ 1.0 \\
    Maximum output / turns & 65,536 tokens / 250 turns \\
    \midrule
    \multicolumn{2}{@{}l}{\textit{Supervised fine-tuning}} \\
    Base models & Qwen3.5-4B; Qwen3.6-27B; Qwen3.6-35B-A3B \\
    Full-corpus training tasks & 112,686 \\
    Parameter updates & Language model unfrozen; vision and aligner frozen \\
    Epochs & 4 \\
    Sequence length & 256K tokens \\
    Global batch size & 32 packed sequences \\
    Peak learning rate & $3\times10^{-5}$ \\
    Optimizer & Distributed; $\beta_1=0.9$, $\beta_2=0.95$ \\
    Schedule / warmup & Cosine to $2\times10^{-6}$ / 10 iterations \\
    Weight decay / clipping & 0.01 / 1.0 \\
    Precision & BF16 \\
    Packing & 262,144 tokens; padding-free; overlength rows removed \\
    \bottomrule
  \end{tabularx}
\end{table}

\paragraph{Probe and teacher execution settings.}
Every probe execution, including re-diagnosis after repair, starts in a clean
sandbox with the current visible request and resources. The probe runs in the
Claude Code harness with a 1M-token context window and no context compaction.
It cannot
access internal construction requirements, the Semantic Workflow Graph, the construction plan,
or outputs and reports from previous executions. Teacher executions share this
task-information access boundary and use a 256K-token context window. We
collect trajectories for 94,077 tasks with the GDPval-AA Stirrup
harness~\citep{stirrup} and for the remaining 18,609 with Claude
Code~\citep{claudecode}. Stirrup runs use the GDPval-AA tools and
official context-compaction mechanism\footnote{\url{https://artificialanalysis.ai/methodology/intelligence-benchmarking\#gdpval-aa}},
whereas Claude Code
runs use its native tools and a compaction threshold of 0.7.
Their prompts differ: the teacher attempts the task, while the probe also reports
environmental issues encountered while solving.

\paragraph{Lightweight rollout filtering.}
Every initial teacher rollout undergoes programmatic validity checks. We
exclude rollouts that contain repeated reasoning spans or repeated tool calls,
fail to produce a deliverable, or produce deliverables below a minimum content
threshold. These corpus-wide checks target degenerate or incomplete
executions; they do not provide task-specific judgments of deliverable
correctness.

\paragraph{Selective trajectory rejection sampling.}
In addition to the corpus-wide filters, we select the longest 5\% of initial
teacher rollouts for rejection
sampling using the 95th percentile of total interaction-turn counts.
Length is measured over the complete
rollout rather than individual training sequences, before pairwise
deliverable comparison. Initial rollout length serves as a heuristic for
allocating additional deliverable-quality checks.

For each selected task, the pairwise judge reads the user request, reference
artifacts, and both agents' deliverables, without access to the internal
construction requirements. A DeepSeek-V4-Flash-0731 baseline deliverable is
generated once per task and held fixed across comparisons. A teacher win
retains the corresponding rollout. If the initial teacher attempt loses
or ties, we allow up to three additional attempts (four teacher attempts in
total), stopping at the first win. Each retry starts in a clean sandbox with
the same task inputs and generation settings, without access to previous
attempts or judge feedback. A task contributes no training episode if none
of its attempts wins; that task and its workspace are excluded from the final
corpus. Tasks outside the selected subset retain their initial teacher episodes
after lightweight filtering.

\begin{figure}[t]
  \centering
  \includegraphics[width=0.88\linewidth]{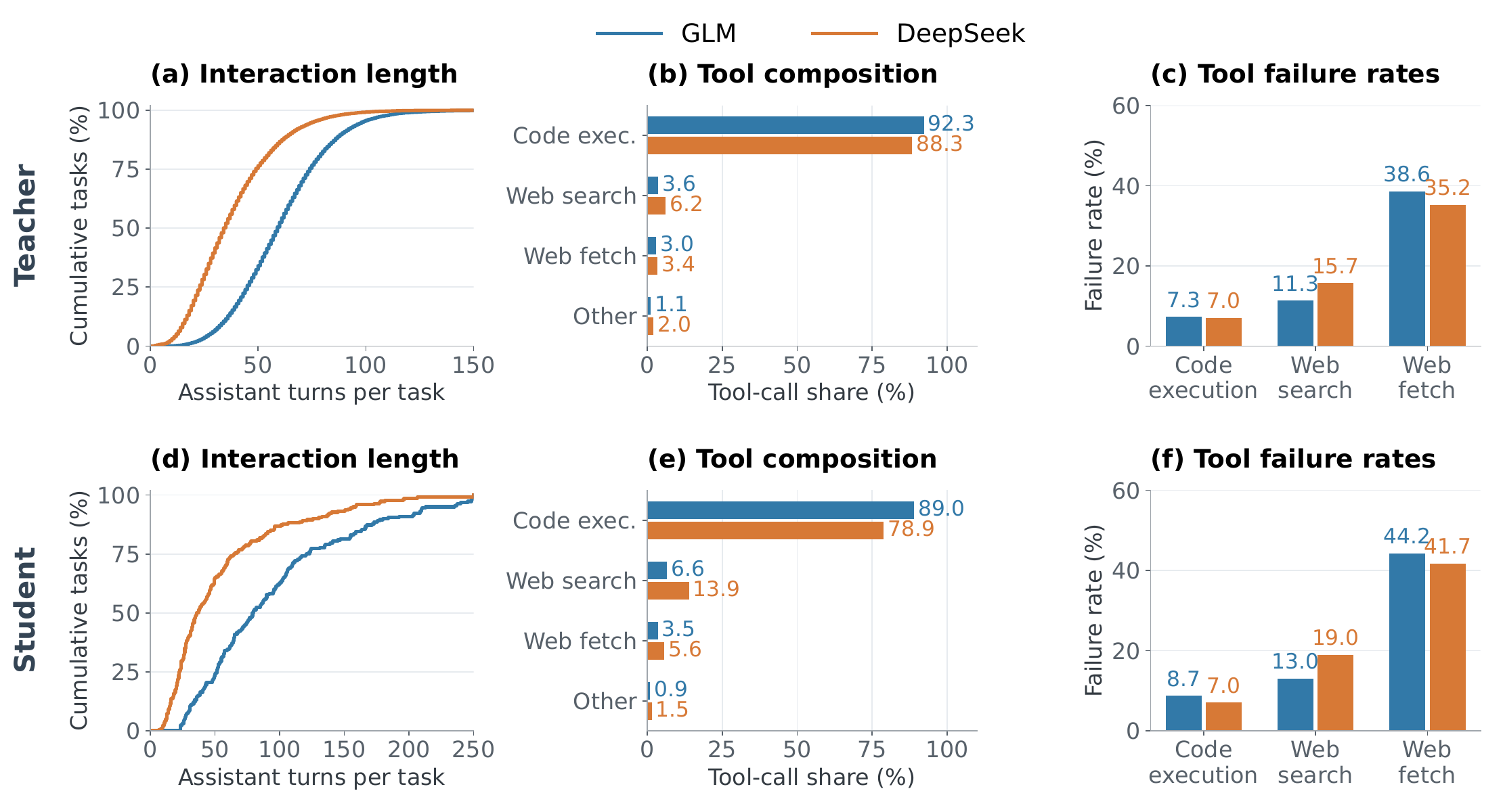}
  \caption{Teacher and student tool-use behavior. The top row uses teacher
  trajectories on the matched 30K-task training subset; the bottom row uses student
  rollouts on the same 220 GDPval evaluation tasks.
  \textbf{(a,d)} Empirical cumulative distributions of task-level mean assistant
  turns. \textbf{(b,e)} Tool-call composition, obtained by summing per-task mean
  call counts and normalizing across tool types; Other includes remaining tools.
  \textbf{(c,f)} Task-macro failure rates. For student results, the three evaluation
  rollouts are averaged within each task for turns and call composition; for
  failure rates, failures and calls from the three rollouts are pooled within a
  task before averaging rates over tasks invoking that tool.
  GLM denotes GLM-5.3 in the teacher row and its trained student in the student
  row; DeepSeek analogously denotes DeepSeek-V4-Flash-0731 and its student.}
  \label{fig:teacher-sensitivity}
\end{figure}

\begin{table}[t]
  \centering
  \caption{Trajectory and student-behavior statistics for the 30K-task teacher comparison. Student behavioral statistics are
  task-level means over the same 220 GDPval tasks after averaging three
  independent evaluation rollouts within each task. All token counts use the
  Qwen3.6-35B-A3B tokenizer. GLM denotes GLM-5.3; DeepSeek denotes
  DeepSeek-V4-Flash-0731.}
  \label{tab:teacher-sensitivity}
  \small
  \setlength{\tabcolsep}{3.5pt}
  \begin{tabular}{@{}lrr@{}}
    \toprule
    \textbf{Metric} & \textbf{GLM source} & \textbf{DeepSeek source} \\
    \midrule
    \multicolumn{3}{@{}l}{\textit{Teacher behavior on the 30K subset}} \\
    Mean assistant turns          & 60.69 & 37.97 \\
    Mean tool calls               & 65.76 & 44.92 \\
    Mean reasoning tokens         & 48,127 & 48,368 \\
    \midrule
    \multicolumn{3}{@{}l}{\textit{Student evaluation on 220 shared tasks}} \\
    GDPval-AA Elo                 & 1483 & 1248 \\
    Bootstrap 95\% CI            & $[1455,1510]$ & $[1220,1274]$ \\
    Mean assistant turns          & 99.81 & 53.81 \\
    Mean tool calls               & 110.40 & 65.95 \\
    Mean reasoning tokens         & 79,076 & 61,723 \\
    \bottomrule
  \end{tabular}
\end{table}

\paragraph{Sequence preprocessing and supervision.}
Selected teacher rollouts are converted into training sequences. A task can
contribute multiple sequences when its rollout is compacted. Sequences
exceeding 262,144 tokens are removed before packing; the remaining sequences
form $\mathcal S$. The reported corpus and subset sizes count distinct tasks
represented by at least one retained sequence rather than the sequences
themselves.
We use masked autoregressive supervision:
\begin{equation}
 \begin{aligned}
 \mathcal J_{\mathrm{SFT}}(\theta)
 &=-\frac{1}{\sum_{s\in\mathcal S}|M_s|}\\
 &\quad\cdot\sum_{s\in\mathcal S}\sum_{\ell\in M_s}
 \log \pi_\theta(a_{s,\ell}\mid H_{s,<\ell}),
 \end{aligned}
 \label{eq:sft}
\end{equation}
where $M_s$ denotes supervised assistant-token positions in sequence $s$, $a_{s,\ell}$ the target token, and $H_{s,<\ell}$ the preceding context within that sequence, including actions and observations.
We identify unsuccessful tool calls from execution-failure indicators in their
tool results, such as a traceback or a nonzero exit code. Only the token span
of each failed tool call is removed from $M_s$; reasoning and ordinary text
in the same assistant turn remain supervised. The failed call and its tool
result remain in the history so that subsequent recovery actions retain
their execution context.

\subsection{Teacher-to-Student Behavior Analysis}
\label{app:teacher-sensitivity}

For the GLM-5.3 condition, we reuse the Qwen3.6-35B-A3B checkpoint trained on the 30K
subset in the scaling study (Figure~\ref{fig:task-scaling}). 
To construct a task-matched DeepSeek condition, we regenerate trajectories for the same WorkFactory tasks with
DeepSeek-V4-Flash-0731 using the Stirrup harness, then train the same base
model with the shared supervised fine-tuning configuration. 
Figure~\ref{fig:teacher-sensitivity} visualizes the teacher-to-student behavior
comparison, and Table~\ref{tab:teacher-sensitivity} summarizes the mean trajectory statistics and student Elo estimates with bootstrap
95\% confidence intervals.

Students trained on trajectories from either teacher substantially outperform
the 879-Elo base model, showing that the gains from WorkFactory training are not exclusive to GLM-5.3 supervision.
On the matched training tasks, GLM-5.3 uses more assistant
turns and tool calls than DeepSeek-V4-Flash-0731 despite comparable
reasoning-token counts. 
On the shared evaluation tasks, the GLM-trained student likewise uses more assistant turns and tool calls than the DeepSeek-trained student.
Figure~\ref{fig:teacher-sensitivity}
shows the interaction-length distributions, tool-call composition, and
tool-failure rates for both teachers and their students.

\begin{figure}[t]
  \centering
  \includegraphics[width=0.92\linewidth]{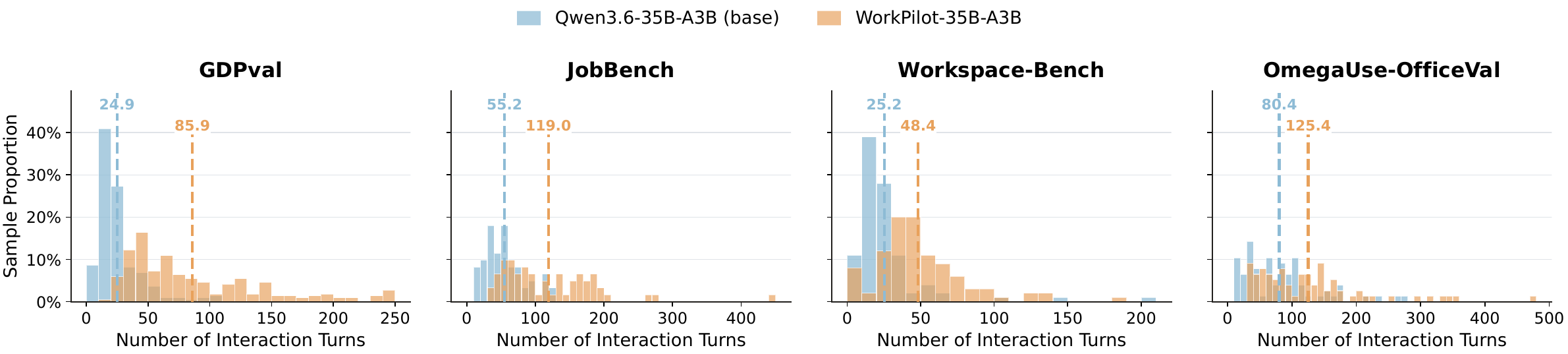}
  \caption{Distribution of interaction turns for Qwen3.6-35B-A3B and
  WorkPilot-35B-A3B across four professional-work benchmarks. Histograms show
  within-model sample proportions; dashed lines and numeric labels denote mean
  interaction lengths.}
  \label{fig:turns-four-benchmarks}
\end{figure}

\section{Evaluation Details}
\label{app:experimental-details}

Unless otherwise noted, all locally run evaluations use three independent
rollouts for each model--task pair.

\paragraph{Text-only agent evaluation.}
GLM-5.3, the teacher model, supports text input only. WorkPilot training
therefore uses text-only teacher trajectories and updates only the language
model, leaving the native vision and alignment components of the Qwen
backbones frozen. We evaluate WorkPilot and its matched Qwen base models under
the same text-only setting: the harnesses do not pass visual inputs to the
task-solving models. This restriction does not apply to evaluation judges,
which may inspect rendered documents and native media when assessing the
resulting deliverables.

\paragraph{Cross-benchmark interaction lengths.}
Figure~\ref{fig:turns-four-benchmarks} compares the distributions of
interaction length for Qwen3.6-35B-A3B and WorkPilot-35B-A3B across the four
evaluation benchmarks. WorkPilot executes longer trajectories on all four
benchmarks, suggesting that WorkFactory training enables the model to sustain
more extended workflows rather than terminating early. This cross-sectional
comparison complements Figure~\ref{fig:task-scaling}(b): along the WorkPilot
scaling trajectory, interaction length rises sharply at small data scales and
then declines with additional training, although the final model still uses
more turns than the untrained base model.

\subsection{Two-Anchor Elo Estimation}
\label{app:two_anchor_elo}

We estimate a target checkpoint against two fixed anchors, GLM-5.3 ($A$) and
DeepSeek-V4-Flash-0731 ($B$), and map its fitted Bradley--Terry (BT)
utility~\citep{bradley1952rank} to the official GDPval-AA v2.1 scale. The
official GDPval-AA v2.1 ratings were retrieved on September 26, 2026, using the
configuration shown on the leaderboard on that date~\citep{gdpvalaa}. The
evaluation covers all 220 tasks, spanning nine sectors and 44 occupations. A
Gemini 3.1 Pro Preview judge~\citep{gemini31propreview} receives the original
task instruction and associated reference materials, together with two
anonymized submissions for the same task. Artifact evidence includes extracted
file text, rendered document images, and native media where applicable. No
task-specific rubrics or rubric-derived scores are provided. Instead, the
judge assesses each submission against the original task requirements using a
shared pairwise evaluation prompt that emphasizes correctness, completeness,
usability, professional quality, and faithfulness to the requested
deliverables. It returns a preference (A, B, or tie) together with concise,
evidence-grounded reasons; Appendix~\ref{app:pairwise_judge_prompt} provides
the prompt. The presentation order is balanced, and outcomes are converted to
the target model's perspective: wins, ties, and losses receive credit
$y\in\{1,0.5,0\}$.

Let $\beta_i$ denote model $i$'s BT utility. Pairwise win probability is
\begin{equation}
  p_{ij}=\sigma(\beta_i-\beta_j), \qquad
  \sigma(x)=(1+e^{-x})^{-1}.
  \label{eq:bt_probability}
\end{equation}
We freeze $\beta_A=0$ and $\beta_B=-1.4645345552$. For target $m$, let $n_a$
denote the number of valid comparisons against anchor $a$, and let $w_a$ be
the accumulated half-win credit. Writing
$q_a(\beta)=\sigma(\beta-\beta_a)$, we maximize
\begin{equation}
  \begin{aligned}
  \ell_m(\beta)={}&\sum_{a\in\{A,B\}}w_a\log q_a(\beta)\\
  &+\sum_{a\in\{A,B\}}(n_a-w_a)\log(1-q_a(\beta)).
  \end{aligned}
  \label{eq:target-likelihood}
\end{equation}
Equivalently, we solve the first-order score equation
\begin{equation}
  \sum_{a\in\{A,B\}}n_a\sigma(\beta-\beta_a)
  =\sum_{a\in\{A,B\}}w_a
\end{equation}
with a bracketed one-dimensional root finder on $[-20,20]$. Comparisons with
unresolved judge errors are excluded rather than imputed. Retry-successful
deliverables are retained; under the strict policy, a completed comparison is
replaced by a target-side loss if the target rollout previously exceeded the
maximum-turn limit. The permissive sensitivity estimate instead retains the
judge's observed verdict. This override is applied only to the target, not to
the fixed anchors.

\begingroup
\setlength{\emergencystretch}{2em}
We map the fitted utility to v2.1 Elo through three frozen calibration nodes:
Qwen3.6-35B-A3B $(-4.4666181420,879)$, Qwen3.8-27B
$(-1.4179100010,1409)$, and GLM-5.3 $(0,1646)$. We use linear interpolation
between adjacent nodes and nearest-segment extrapolation outside their range.
The transformation is monotone and therefore preserves the ordering of fitted
utilities.
\par
\endgroup

For WorkPilot-35B-A3B, we obtain $\widehat\beta=-0.486737$, which maps to
1564.64 Elo and is reported as 1565. Strict and permissive estimates coincide
because no comparison receives a maximum-turn override. To quantify
task-sampling uncertainty, we resample tasks with replacement and retain all
comparisons associated with each sampled task. For each of 1,000 bootstrap
draws, we refit $\beta$ and apply the frozen calibration. The resulting 95\%
percentile interval is $[1539.20,1588.65]$, reported as $[1539,1589]$. This
interval is conditional on the task population, anchors, judge, and calibration
and does not include method-form error.

Table~\ref{tab:local-elo-ci} reports the same task-bootstrap uncertainty for
all local Elo estimates in Table~\ref{tab:workfactory-benchmarks}. Point
estimates and interval endpoints are rounded to the nearest Elo point.

\begin{table}[t]
  \centering
  \caption{Task-bootstrap 95\% confidence intervals for the locally estimated GDPval-AA v2.1 Elo ratings in Table~\ref{tab:workfactory-benchmarks}.}
  \label{tab:local-elo-ci}
  \small
  \setlength{\tabcolsep}{5pt}
  \renewcommand{\arraystretch}{1.08}
  \begin{tabular}{@{}lrr@{}}
    \toprule
    \textbf{Model} & \textbf{Local Elo} & \textbf{Bootstrap 95\% CI} \\
    \midrule
    Nex-N2.5-mini     & 1127 & $[1085,1163]$ \\
    Qwen3.5-4B        &  661 & $[363,792]$ \\
    WorkPilot-4B      & 1375 & $[1344,1405]$ \\
    Qwen3.6-27B       &  975 & $[900,1038]$ \\
    WorkPilot-35B-A3B & 1565 & $[1539,1589]$ \\
    WorkPilot-27B     & 1612 & $[1594,1631]$ \\
    \bottomrule
  \end{tabular}
\end{table}

The wider interval for Qwen3.5-4B reflects weak identification in the lower
tail: because both judging anchors are substantially stronger, few target-side
wins are available to locate its utility precisely.

\begin{figure}[t]
  \centering
  \includegraphics[width=0.66\linewidth]{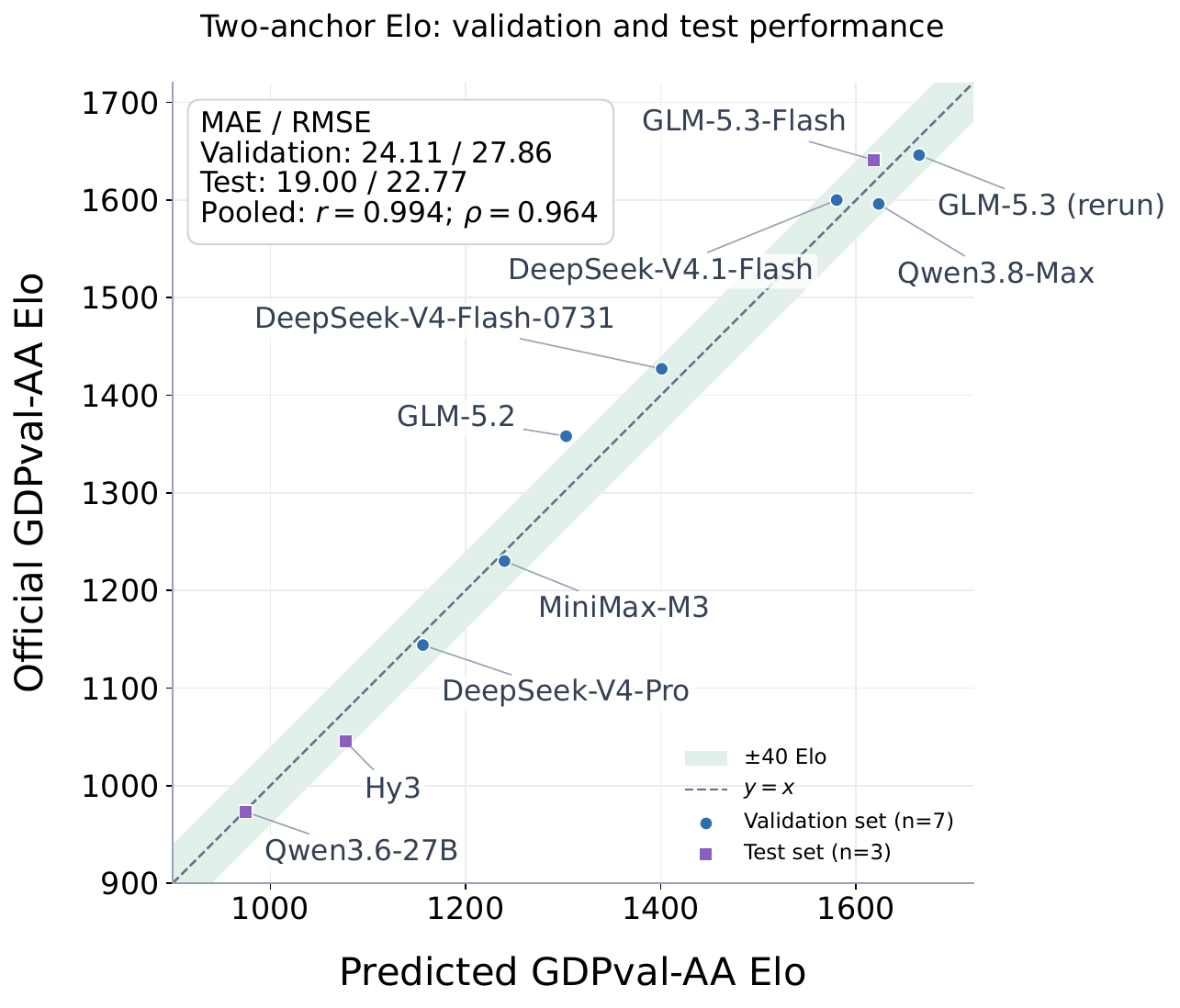}
  \caption{Predicted versus official GDPval-AA Elo. Blue circles show the
  seven-model validation set used to determine the estimator design and
  calibration; purple squares show the three-model test set evaluated after
  the procedure was frozen. Predicted Elo is on the horizontal axis and
  official Elo is on the vertical axis. The dashed line denotes equality and
  the shaded region denotes $\pm40$ Elo. The independently evaluated GLM-5.3
  rerun is distinct from the GLM-5.3 calibration node.}
  \label{fig:elo_validation}
\end{figure}

\paragraph{Empirical validation.}
We use seven previously evaluated models that are not calibration knots as a
validation set to determine the estimator design and calibration. On this set,
the frozen estimator has an MAE of 24.11 Elo, an RMSE of 27.86 Elo, a bias of
$-4.71$ Elo, and a maximum absolute residual of 54.96 Elo. After fixing the
complete estimation procedure, we evaluate it on three subsequently available
models that were not used to determine the calibration. On this test set, the
method achieves an MAE of 19.00 Elo, an RMSE of 22.77 Elo, a bias of $+3.89$
Elo, and a maximum absolute error of 32.20 Elo. Across both sets, predicted and
official Elo are strongly aligned (Pearson $r=0.994$; Spearman $\rho=0.964$),
and nine of ten models lie within $\pm40$ Elo of their official ratings
(Figure~\ref{fig:elo_validation}). Removing eight tasks requiring multimodal
evidence leaves model ordering unchanged and changes the combined-panel RMSE
only from 26.43 to 27.56 Elo. These empirical errors are reported separately
from the task-bootstrap interval because they measure different sources of
uncertainty.

\subsection{Pairwise Judge Prompt}
\label{app:pairwise_judge_prompt}

\begin{promptbox}
You are the pairwise evaluator for GDPVal-AA. Determine which blinded submission better fulfills the original professional task.

Evaluate both submissions using all available evidence from the initial task, reference files, extracted file text, rendered document images, native media, and submission files. Consider correctness, completeness, usability, professional quality, and faithfulness to requested deliverables.

Rules:
1. The identities of the systems are intentionally hidden. Never infer model identity from style, filenames, metadata, or self-description, and never favor Submission A or B because of its position.
2. Inspect both submissions independently before comparing them. Missing, empty, corrupt, or unusable deliverables are substantive failures.
3. For factual or pointable claims, rely on evidence actually present in the submitted artifacts. A submission's claim that it is correct is not proof.
4. For visual deliverables, use rendered images as well as extracted text. When visible rendering conflicts with extraction, trust the visible artifact.
5. Use tie only when the submissions are genuinely equivalent overall or the available evidence cannot support a defensible preference. Do not use a tie merely because each submission has different strengths.
6. Check all material requirements in the original task, but keep the response concise. Return only the most decision-relevant comparisons.

Return exactly one JSON object containing (i) a winner in {A, B, tie}, (ii) a concise evidence-grounded overall explanation, and (iii) at most ten decisive comparisons. Each decisive comparison contains the relevant criterion, which submission is better, and a short specific evidence pointer for each submission. Return valid JSON without Markdown or surrounding text, and do not include model names, guessed identities, Elo ratings, pointwise scores, or hidden metadata.
\end{promptbox}

\subsection{GDPval-AA Inference-Cost Accounting}
\label{app:cost-performance}

\begin{table}[H]
  \centering
  \caption{Complete pricing assignment for the GDPval-AA cost--performance analysis. Input, cache-read, and output prices are USD per one million tokens, using the October 12, 2026 snapshot.}
  \label{tab:cost-price-mappings}
  \fontsize{6.5pt}{7pt}\selectfont
  \setlength{\tabcolsep}{1.8pt}
  \renewcommand{\arraystretch}{0.94}
  \begin{tabularx}{\textwidth}{@{}p{0.15\textwidth}p{0.22\textwidth}rrrY@{}}
    \toprule
    \textbf{Pricing basis} & \textbf{OpenRouter endpoint / provider} &
    \textbf{Input} & \textbf{Cache} & \textbf{Output} &
    \textbf{Models assigned to this basis and mapping rule} \\
    \midrule
    WorkPilot-4B (scaled Nex-N2.5-mini) & \nolinkurl{nex-agi/nex-n2.5-mini}; Nex AGI &
      0.00625 & 0.000625 & 0.025 &
      WorkPilot-4B; each Nex rate is multiplied by $1/4$. \\
    Nex-N2.5-mini & \nolinkurl{nex-agi/nex-n2.5-mini}; Nex AGI &
      0.025 & 0.0025 & 0.10 &
      WorkPilot-35B-A3B, Nex-N2.5-mini, Qwen3.6-35B-A3B, and Apodex-1.1-mini; same-size alignment. \\
    Qwen3.8-27B & \nolinkurl{qwen/qwen3.8-27b}; Wafer &
      0.04 & 0.0198 & 2.30 &
      WorkPilot-27B, Qwen3.8-27B, and Qwen3.6-27B; same-size alignment. \\
    Qwen3.8-Max-0902 & \nolinkurl{qwen/qwen3.8-max-0902}; Alibaba Cloud Int. &
      2.00 & 0.25 & 6.00 &
      Qwen3.8-Max; current endpoint used as a version proxy. \\
    GLM-5.3 & \nolinkurl{z-ai/glm-5.3}; Wafer &
      0.07 & 0.05 & 4.80 & GLM-5.3; direct endpoint. \\
    GLM-5.2 & \nolinkurl{z-ai/glm-5.2}; Wafer &
      0.06 & 0.059 & 7.00 & GLM-5.2; direct endpoint. \\
    MiniMax-M3 & \nolinkurl{minimax/minimax-m3}; MiniMax &
      0.30 & 0.06 & 1.20 & MiniMax-M3; direct endpoint. \\
    Claude Opus 4.7 & \nolinkurl{anthropic/claude-opus-4.7}; Anthropic &
      5.00 & 0.50 & 25.00 & Claude Opus 4.7; direct Anthropic endpoint. \\
    Claude Opus 4.8 & \nolinkurl{anthropic/claude-opus-4.8}; Anthropic &
      5.00 & 0.50 & 25.00 & Claude Opus 4.8; direct Anthropic endpoint. \\
    Kimi-K3 & \nolinkurl{moonshotai/kimi-k3}; Wafer &
      1.95 & 0.30 & 30.00 & Kimi-K3; direct endpoint. \\
    DeepSeek-V4.1-Flash & \nolinkurl{deepseek/deepseek-v4.1-flash}; Wafer &
      0.035 & 0.034 & 1.60 & DeepSeek-V4.1-Flash; direct endpoint. \\
    DeepSeek-V4-Flash-0731 & \nolinkurl{deepseek/deepseek-v4-flash-0731}; Wafer &
      0.13 & 0.12 & 0.23 & DeepSeek-V4-Flash-0731; direct endpoint. \\
    DeepSeek-V4-Pro-0813 & \nolinkurl{deepseek/deepseek-v4-pro-0813}; Wafer &
      0.36 & 0.219 & 5.00 &
      DeepSeek-V4-Pro-0813 and DeepSeek-V4-Pro; Preview version aligned to 0813. \\
    GPT-5.6 Sol & \nolinkurl{openai/gpt-5.6-sol}; OpenAI &
      4.00 & 0.40 & 20.00 & GPT-5.6 Sol; direct endpoint. \\
    Hy3 & \nolinkurl{tencent/hy3}; Tencent Cloud &
      0.132 & 0.033 & 0.528 & HY3; direct endpoint. \\
    Hy4 Preview & \nolinkurl{tencent/hy4-preview}; Tencent Cloud &
      0.834 & 0.042 & 2.501 & HY4 Preview; direct endpoint. \\
    \bottomrule
  \end{tabularx}
\end{table}

We estimate model-inference cost per GDPval-AA task from the input and output
tokens recorded in local evaluation logs. For trial $r$ of task $j$, with
input-token count $T^{(j,r)}_{\mathrm{in}}$, output-token count
$T^{(j,r)}_{\mathrm{out}}$, cache-hit rate $h$, and prices
$P_{\mathrm{in}}$, $P_{\mathrm{cache}}$, and $P_{\mathrm{out}}$ in USD per
million tokens, we compute
\begin{equation}
c_{j,r} = \frac{T^{(j,r)}_{\mathrm{in}}}{10^6}
    \bigl((1-h)P_{\mathrm{in}}+hP_{\mathrm{cache}}\bigr)
  + \frac{T^{(j,r)}_{\mathrm{out}}}{10^6}P_{\mathrm{out}}.
\label{eq:trial-inference-cost}
\end{equation}
For each task, we first average over its completed trials $R_j$, and then
average equally over the included tasks $J$:
\begin{equation}
C_{\mathrm{task}}
= \frac{1}{|J|}\sum_{j\in J}
  \left(\frac{1}{|R_j|}\sum_{r\in R_j}c_{j,r}\right).
\label{eq:task-inference-cost}
\end{equation}
We set $h=0.992$ for all models. Reasoning tokens are already included in
output-token totals and are not counted again. Only trials without recorded
execution exceptions are included, and tasks with no completed trial are
excluded. This two-stage averaging gives each included task equal weight even
when the number of completed trials differs. Because cache-write usage is not
consistently exposed in the logs, we do not add a separate cache-write charge.

We use provider-specific OpenRouter\footnote{\url{https://openrouter.ai}} prices recorded on October 12,
2026. Where an evaluated checkpoint lacks a directly
matching endpoint, we use an explicit version proxy or a common price for the
corresponding model-size class. Table~\ref{tab:cost-price-mappings} records the
complete mapping used for every model in Figure~\ref{fig:cost-pareto}; prices
are in USD per one million tokens.
These values are standardized estimates rather than provider invoices. They
exclude judges, verifiers, tool and web-search fees, sandbox compute, storage,
and orchestration overhead. Provider prices may also vary by endpoint and over
time; same-size alignments intentionally replace model-specific commercial
prices so that differences within a class primarily reflect observed token
usage.

% \begin{PromptBox}[p:swr_prompt]{Prompt for SWR}
% \begin{lstlisting}
% System Prompt:

% \end{lstlisting}
% \end{PromptBox}

\end{document}